\documentclass[times, review, 10pt]{elsarticle}

\usepackage{amssymb}
\usepackage{amsmath}
\usepackage{hyperref}
\usepackage{dsfont}
\usepackage{graphicx}
\usepackage{tikz}
\usepackage{multirow}
\usepackage{subcaption}
\usepackage{float}
\usepackage{graphicx}
\usepackage{mathtools}
\usepackage{microtype}
\usepackage{multicol}
\usepackage{geometry}
\usepackage{tabularx}
\usepackage{tikz}
\usepackage{url}
\allowdisplaybreaks

\usepackage{algorithm}
\usepackage{placeins}   
\usepackage{algpseudocode} 

\journal{Pattern Recognition}

\begin{document}

\begin{frontmatter}



\title{A statistical method for crack pre-detection in 3D concrete images}


\author[label1]{Vitalii Makogin\corref{cor2}} 
\author[label1]{Duc Nguyen\corref{cor1}}%
\ead{tran-1.nguyen@uni-ulm.de} 
\author[label1]{Evgeny Spodarev} 
\ead{evgeny.spodarev@uni-ulm.de} 

\cortext[cor2]{Dedicated to the memory of Dr. Vitalii Makogin (12.12.1987 - 08.05.2024)}
\cortext[cor1]{Corresponding author}

\affiliation[label1]{organization={Institute of Stochastics, Ulm University},
            addressline={\\Helmholtzstraße 16}, 
            city={Ulm},
            postcode={89081}, 
            state={Baden-Württemberg},
            country={Germany}}

\begin{abstract}

In practical applications, effectively segmenting cracks in large-scale computed tomography (CT) images holds significant importance for understanding the structural integrity of materials. Classical image-processing techniques and modern deep-learning models both face substantial computational challenges when applied directly to high resolution big data volumes. This paper introduces a statistical framework for crack pre-localization, whose purpose is not to replace or compete with segmentation networks, but to identify, with controlled error rates, the regions of a 3D CT image that are most likely to contain cracks. The method combines a simple Hessian-based filter, geometric descriptors computed on a regular spatial partition, and a spatial multiple testing procedure to detect anomalous regions while relying only on minimal calibration data, rather than large annotated datasets. Experiments on semi-synthetic and real 3D CT scans demonstrate that the proposed approach reliably highlights regions likely to contain cracks while preserving linear computational complexity. By restricting subsequent high resolution segmentation to these localized regions, deep-learning models can be trained and operate more efficiently, reducing both training runtime as well as resource consumption. The framework thus offers a practical and interpretable preprocessing step for large-scale CT inspection pipelines.
\end{abstract}



\begin{keyword}


CUSUM \sep multiple hypotheses testing \sep Hessian-based filter \sep crack detection \sep classification.
\end{keyword}

\end{frontmatter}



\section{Introduction}
\label{sec1}

Concrete serves as the foundational material for various structures, including buildings and bridges, underscoring the importance of ensuring its quality, durability, and mechanical stability. To gain a better understanding of the mechanisms and causes of cracks, it is essential to subject concrete specimens to stress tests, which is valuable to prevent the occurrence of emergency conditions. Experimental studies under cyclic loading have clarified how cracks initiate and evolve in concrete, providing a foundation for understanding failure mechanisms \cite{guo2023,jia2022experimental} while recent reviews emphasize the need for systematic and reproducible methodologies for their reliable detection and quantification \cite{kaveh2024recent,lee2024parametric,flah2020classification}. 

The use of computed tomography (CT) imaging offers a powerful means to study crack propagation in concrete with high spatial and volumetric resolution. In CT images of concrete, cracks typically appear as regions of lower intensity as compared to the background material. 

This characteristic suggests the use of segmentation methods involving the application of a global threshold or pixel/voxel-wise analysis \cite{ogawa2019}. Geometric descriptors such as circularity and roundness have further been shown to distinguish cracks from pores in X-ray CT data \cite{he2025quantitative,morozova2023frost}, demonstrating the potential of shape-based characterization. However, a modern CT scanner \cite{salamon2025gulliver} generates ultra-large volumetric datasets, often containing millions to billions of voxels with dimensions up to $10000 \times 10000 \times 2000$, posing significant computational challenges for traditional image analysis methods and creating practical bottlenecks for routine inspection workflows.

Classical image segmentation methods for crack analysis can be classified into distinct categories. First, global thresholding based on gray-level statistics and assumptions about intensity distributions \cite{ogawa2019,morozova2022visualization,avendano2024image} provides a straightforward but often noise-sensitive approach. Second, edge-detection techniques based on the Hessian matrix and its eigenvalue analysis, such as the Frangi filter and Sheet filter \cite{frangi98, sato2000tissue}, effectively identify flat-like or vessel-like structures within materials. Region-growing methods such as percolation-based filter \cite{Yamaguchi2010, ehrig2011comparison} provide an alternative that identifies cracks through voxel connectivity. A systematic comparison on semi-synthetic concrete volumes \cite{barisin2022,zawad2021comparative} shows that these classical approaches perform competitively when tuned appropriately. While these classical methods can effectively identify crack structures, they face significant limitations including sensitivity to noise, high computational costs in three-dimensional contexts, and limited statistical control over false positive rates. 

Deep learning-based crack detection in concrete has developed along complementary lines, with 2D surface analysis remaining widely used and 3D volumetric modeling becoming increasingly prominent. In the 2D setting, YOLO-based detection-segmentation models have become standard for thin surface defects, ranging from cracks in civil concrete structures to crack-like flaws on metallic surfaces \cite{qi2025crack, zhou2025mpa}.
Lightweight YOLO-based architectures that combine pruning and knowledge distillation further reduce inference cost while maintaining segmentation accuracy on concrete structures \cite{xu2025crack,zhu2024lightweight}, and encoder–decoder models such as Cracklab \cite{yu2022cracklab} and STRNet \cite{kang2022efficient} achieve accurate segmentation of thin cracks on heterogeneous surfaces. Apart from architectural advances, several works shift attention to training-data optimization. Using multi-stage label refinement, synthetic crack data and diffusion-based resolution reduce noise effects in real-world concrete scans \cite{li2025high,xie2025versatile,xia2026image}.

In three-dimensional volumetric space, cracks have been modeled as minimum-weight surfaces in 3D Voronoi diagrams, enabling the synthesis of realistic crack geometries and associated morphological descriptors directly in voxel space \cite{jung2023crack}. This approach underlies the semi-synthetic VoroCrack3D dataset, where such crack surfaces are embedded into $\mu$CT volumes of different concrete types to provide high-quality 3D ground truth for training and evaluation \cite{JUNG2024110474}. Based on this dataset, 3D U-Net and 3D Feature Pyramid Network architectures have demonstrated that deep networks can segment complex, branching crack networks in heterogeneous microstructures, but they still struggle with extremely thin or low-contrast internal cracks and do not provide formal control of false detections in large 3D volumes \cite{nowacka2024deep}. These limitations highlight the need for methods that combine volumetric crack pre-detection with statistically principled error control.

Motivated by these challenges, we introduce a statistical framework for crack pre-localization in large three-dimensional concrete CT images. Instead of employing deep learning segmentation networks or classical techniques that demand large training sets and sensitive parameter calibration, our aim is to identify regions, under false positive rate control, that are most likely to contain cracks. From a statistical point of view, our method provides a confidence inference region around the spatial crack pattern, in analogy to the confidence regions used for functional parameters in functional data analysis \cite{ramsay2005fda,liebl2023fast}. 
We first apply a simple Hessian-based filter to preserve crack structures and then compute geometric descriptors on a regular spatial partition, which yields a crack-sensitive random field defined on cubes. This field is evaluated over a family of overlapping scanning windows and analysed using a spatial multiple testing procedure inspired by the methodology in \cite{cai2021}, providing control of the expected proportion of false discoveries in the scanned image. The whole procedure is described in Figure \ref{fig:process_specified}. 
\begin{figure}[htbp]
	\centering
	\begin{tikzpicture}
		\node (step0) at (5,2) [draw, rectangle, text width=4cm, align=center] {3D input image};
		\node (step1) at (0,1) [draw, rectangle, text width=4cm, align=center] {Image binarization};
		\node (step2) at (5,1) [draw, rectangle, text width=4cm, align=center] {Binary image subdivision};
		\node (step3) at (10,1) [draw, rectangle, text width=4cm, align=center] {Geometric \\statistics' extraction};
		\node (step4) at (0,-0.15) [draw, rectangle, text width=4cm, align=center] {Random fields generation};
		\node (step5) at (5,-0.15) [draw, rectangle, text width=4cm, align=center] {Multiple testing};
		\node (step6) at (10,-0.15) [draw, rectangle, text width=4cm, align=center] {Crack pre-detection};
		
		\draw[->, thick] (step0) -- (step1);
		\draw[->, thick] (step1) -- (step2);
		\draw[->, thick] (step2) -- (step3);
		\draw[->, thick] (step3) -- (step4);
		\draw[->, thick] (step4) -- (step5);
		\draw[->, thick] (step5) -- (step6);
	\end{tikzpicture}
	\caption{7-step processing pipeline diagram for crack pre-ocalization in 3D concrete images}
	\label{fig:process_specified}
\end{figure}

When applied to both semi-synthetic datasets and real CT volumes, the procedure robustly highlights crack regions using only minimal calibration, with inexpensive computational requirements. Therefore, it can be used to restrict subsequent segmentation, including 3D U-Net models trained on semi-synthetic images, to a small fraction of the volume, thereby reducing computation while preserving the ability to detect relevant crack structures. The reduced spatial domain also indicates the potential for lowering the training burden of deep learning models, since a smaller amount of data would need to be processed during training.

The rest of the paper is organized as follows. Section \ref{sec:crack} introduces the Hessian-based preprocessing methods and the construction of geometric statistics on spatial partitions, which together form the basis of the proposed crack pre-localization framework. Section \ref{sec:anomaly} formulates crack localization as a change-point problem and presents the spatial multiple testing procedure used for anomaly detection. Section \ref{subsec:numeric} reports the numerical evaluation on semi-synthetic and real 3D CT volumes. To support reproducibility, the full implementation of the proposed method,
including preprocessing steps and numerical experiments, is publicly available at
\url{https://github.com/DucQD3/Crack_PreDetect}.
Section \ref{sec:discussion} discusses the practical implications, strengths, and limitations of the method, and Section \ref{sec:conl} concludes the paper and outlines directions for future work.

\section{Crack segmentation} \label{sec:crack}
In this section, we present several classical methods proposed in \cite{frangi98,sato2000tissue} including Frangi filter and Sheet filter. The common idea for these methods is to locally compute the Hessian matrix of an image, which captures its second-order partial derivatives, providing a quantitative representation of local image structures and intensity variations based on its eigenvalues. This approach possesses both pros and contras, notably in terms of computational expenses. Therefore, a simpler method will be introduced to overcome this crucial difficulty. Let $W$ be a subset of $\mathbb{Z}^3$. The grayscale image $I = \{I(p) \in [0,1], p \in W\}$ is employed to represent the input data. The smoothed second-order partial derivatives of the input image $I$ are computed as follows:
\begin{equation}
	\tilde{\partial}^2_{p_i p_j} I(p,\sigma) = \frac{\tilde{\partial}^2 I}{\partial p_i \partial p_j}(p,\sigma)
	= \sigma\, I(p) * \frac{\partial^2}{\partial p_i \partial p_j} G(p,\sigma),
	\ i,j = 1,2,3,\; p=(p_1,p_2,p_3)^{T}\in W.
\end{equation}
where $G$ is the 3-dimensional Gaussian kernel $G(p, \sigma) = (2 \pi \sigma^2)^{-3/2} \exp\{-\left\|p \right\|_2^2/(2\sigma^2)\}$ with scale parameter $\sigma >0$ and $\|\cdot\|_2$ the Euclidean norm in $\mathbb{R}^3$. Here $*$ denotes the usual convolution operation.
The Hessian matrix $H(p, \sigma)$ of an image $I$ at a voxel $p \in W$ is given by
$H(p,\sigma)=\bigl(H_{ij}(p,\sigma)\bigr)_{i,j=1}^3$, $H_{ij}(p,\sigma)=\tilde{\partial}^2_{p_i p_j} I(p,\sigma)$.
\subsection{Classical crack segmentation methods} \label{subsec:segment}

The Sheet filter and Frangi filter predominantly rely on the numerical eigenvalues of $\lambda_1(p, \sigma),\lambda_2(p, \sigma)$,$\lambda_3(p, \sigma)$ of $H(p, \sigma)$. They were designed to detect lower dimensional image features such as vessels and cracks. Given the scale parameter $\sigma$ as well as parameters $\delta > 0, \rho \in (0,1]$ at voxel $p$, denote by $S(p,\sigma)$ and $F(p,\sigma)$ the outcomes produced by Sheet filter and Frangi filter. It is worth noting that the quality of the binary images produced by both methods heavily depends on the scale parameter $\sigma$, recommended to be half of the crack width \cite{barisin2022}. If a voxel $p$ belongs to a crack, it is highly probable that both $S(p, \sigma)$ and $F(p, \sigma)$ will exhibit higher values compared to voxels belonging to homogeneous parts of $I$. Here, both adaptive and manual threshold settings are available. 
However, in practice, the anomalies within a real CT image of a material often do not exhibit ideal conditions, such as a constant width of cracks or significantly higher grayscale contrast at the crack edges, which might make the choice of a threshold challenging. To overcome this disadvantage, one can consider an adaptive global thresholding approach: $S^{*}(p,\sigma) = \mathds{1}\!\left\{\,S(p,\sigma)\ge \mu(S_{\sigma}) + 3\,\mathrm{sd}(S_{\sigma})\,\right\}$ and $F^{*}(p,\sigma) = \mathds{1}\!\left\{\,F(p,\sigma)\ge \mu(F_{\sigma}) + 3\,\mathrm{sd}(F_{\sigma})\,\right\},  p\in W$ where $\mu(I)$ and $sd(I)$ are the sample mean and the sample standard deviation of all gray values within $I$. As an alternative, a multi-scale approach is proposed, which involves computing outcomes from these filters at various scales and selecting the maximum result. Subsequently, a proper threshold is applied. The multi-scale Frangi filter is defined as:
$
	F_{S}(p)=\max_{\sigma_{\min}\le \sigma \le \sigma_{\max}} F(p,\sigma).
$
where $\sigma_{min}$ and $\sigma_{max}$ are chosen in advance.
Furthermore, with the assistance of the two aforementioned filters, one can proactively identify a set of voxels, denoted as $H$, displaying characteristics indicative of potential crack regions. The Hessian-based percolation methodology, originally developed for 2D images \cite{Yamaguchi2010} and later extended to 3D cases \cite{ehrig2011comparison}, provides an iterative strategy to expand the initially detected regions based on local intensity similarities. Starting from the voxel set $H$, the algorithm progressively aggregates neighboring voxels whose gray values meet predefined similarity criteria, thereby connecting adjacent or slightly disconnected elements that are likely to belong to the same crack structure.
	
The effectiveness of this method largely depends on the quality of the initial set $H$, as it determines how well the subsequent region-growing process captures relevant crack voxels. Although this percolation procedure can improve the continuity of cracks and bridge small gaps, it is computationally expensive and typically serves as a post-processing stage applied to the outputs of crack segmentation methods such, e.g., as the Frangi and Sheet filters. Since the main focus of this study lies in evaluating the efficiency and accuracy of these filter-based approaches, the percolation step was not implemented in the performance experiments.

\subsection{Maximal Hessian Entry filter}\label{subsec:Hessian}

With the challenge posed by the variability in crack widths, careful consideration is necessary for computational runtime and storage requirements. Images exceeding dimensions of $600^3$ present significant computational challenges in both Hessian matrix computation and eigenvalue extraction. In response to this challenge, an alternative methodology is proposed-one that avoids extracting maximal information from the Hessian matrix and instead relies exclusively on its entries. Due to the flat structure of cracks, if a voxel $p$ belongs to a crack, the gray values along a line tangent to the crack surface will exhibit a pattern similar to a concave function. Therefore, across all Hessian matrix elements, i.e., the second-order derivatives along six different directions, there should exist at least one element that is non-negative. Choosing the maximal nonnegative entry allows us to capture any sudden changes at voxels belonging to cracks. For an image  $I = \{I(p) \in [0,1], p \in W \subset \mathbb{Z}^3\}$, with a single value of $\sigma$, the Maximal Hessian Entry filter is defined as follows:
\begin{equation}
	L_{\sigma}(p)=\max_{\,i,j=1,2,3}\bigl(H_{i,j}(p,\sigma),\,0\bigr), \ p\in W.
\end{equation}

We apply the same $3\sigma$-rule for the global threshold, which results in a binary image:
\begin{equation}
	L_{\sigma}^{*}(p)=\mathds{1}\!\left\{\,L_{\sigma}(p)
	\ge \mu(L_{\sigma}) + 3\,\mathrm{sd}(L_{\sigma})\,\right\},
	\ p\in W.
\end{equation}
Let $\sigma \in \mathcal{S}$, then compute
$
	L_{\mathcal{S}}(p)=\max_{\sigma\in\mathcal{S}}\,L_{\sigma}^{*}(p),
	 p\in W.
$
The choice of the discrete set \(\mathcal{S}\) can be predetermined according to the nature of the input image. It is important to note that this approach is implemented based solely on the maximal positive entry of the Hessian matrix and a global threshold, leading to a binary image with noticeable noise. However, as observed in the comparison in Section \ref{subsec:binaryperform}, the method still preserves the essential structure of cracks, which is beneficial for our subsequent geometric analysis. 

\subsection{Performance and computational cost} \label{subsec:binaryperform}

\begin{figure}[ht]
	\centering
	\includegraphics[width=\textwidth]{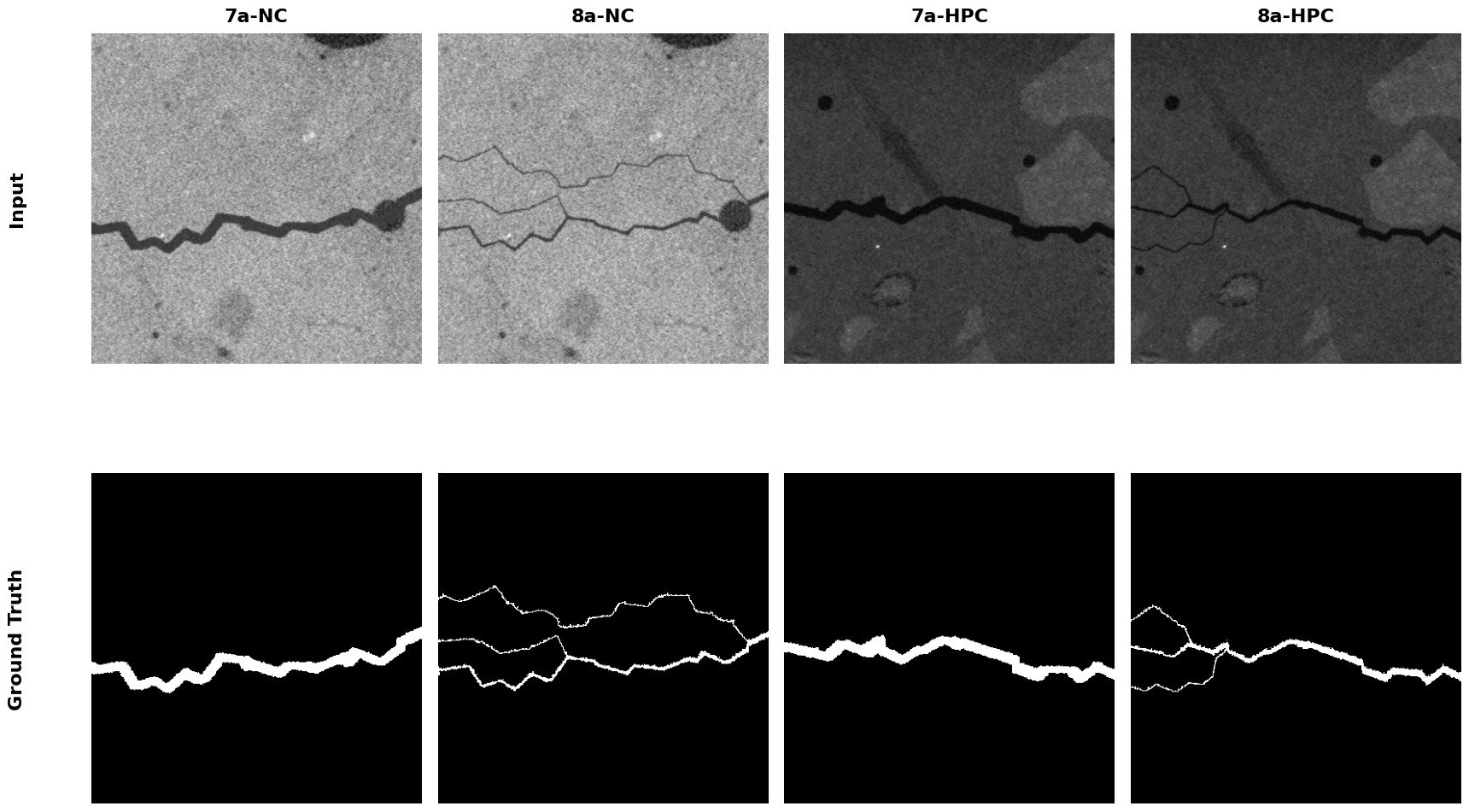}
	\caption{Representative 2D slices from the \textbf{VoroCrack3D} dataset \cite{JUNG2024110474}. 
		The top row shows original semi-synthetic CT images of NC and HPC with different crack patterns (samples 7a and 8a), 
		while the bottom row displays their corresponding ground truth masks. Each volume has a spatial resolution of $400^3$ voxels.}
	\label{fig:syninput}
\end{figure}

		To evaluate the performance of the proposed crack segmentation methods, we applied them to semi-synthetic 3D CT images from the VoroCrack3D dataset \cite{JUNG2024110474}, developed by the Technical University of Kaiserslautern and Fraunhofer ITWM. 		
		In this study, we selected 24 images from the normal concrete (NC) and high-performance concrete (HPC) subfolders of the dataset \cite{JUNG2024110474}. These volumes (4-7) include cracks of fixed widths $w \in \{1, 3, 5, 7\}$ voxels and two different multiscale crack configurations (8-9). To ensure reproducibility, the test set included images corresponding to crack cases (4-9) in their noise-free (a) and low-noise (b) variants from both the NC and HPC subsets of the VoroCrack3D repository. Examples of the semi-synthetic input images with varying crack widths and noise conditions are shown in Figure \ref{fig:syninput}.

	The careful selection of parameters strongly influences the performance of the Frangi and Sheet filters. In this study, the scale parameter $\sigma$ was chosen based on the approximate average crack width $w$ of each test sample.
	For the Frangi filter, we set $\alpha = 0.5$ and $\beta = 0.5$, while for the Sheet filter, we used $\delta = 1$ and $\rho = 0.3$. 	
	For multiscale crack configurations (samples 8a-9b), the cracks exhibit a broad range of local widths, making the selection of a single $\sigma$ value impractical. In these cases, we assigned $\sigma = 2.5$, corresponding to the mean effective width across the structure, to achieve a balanced response for both thin and wide cracks.
	For the Maximal Hessian Entry filter, we employed a multi-scale setting with $\sigma \in \{0.5, 1.5, 2.5, 3.5, 4.5\}$ to ensure sensitivity across the full range of observed crack widths.

\begin{figure}[ht]
	\centering
	\includegraphics[width=\textwidth]{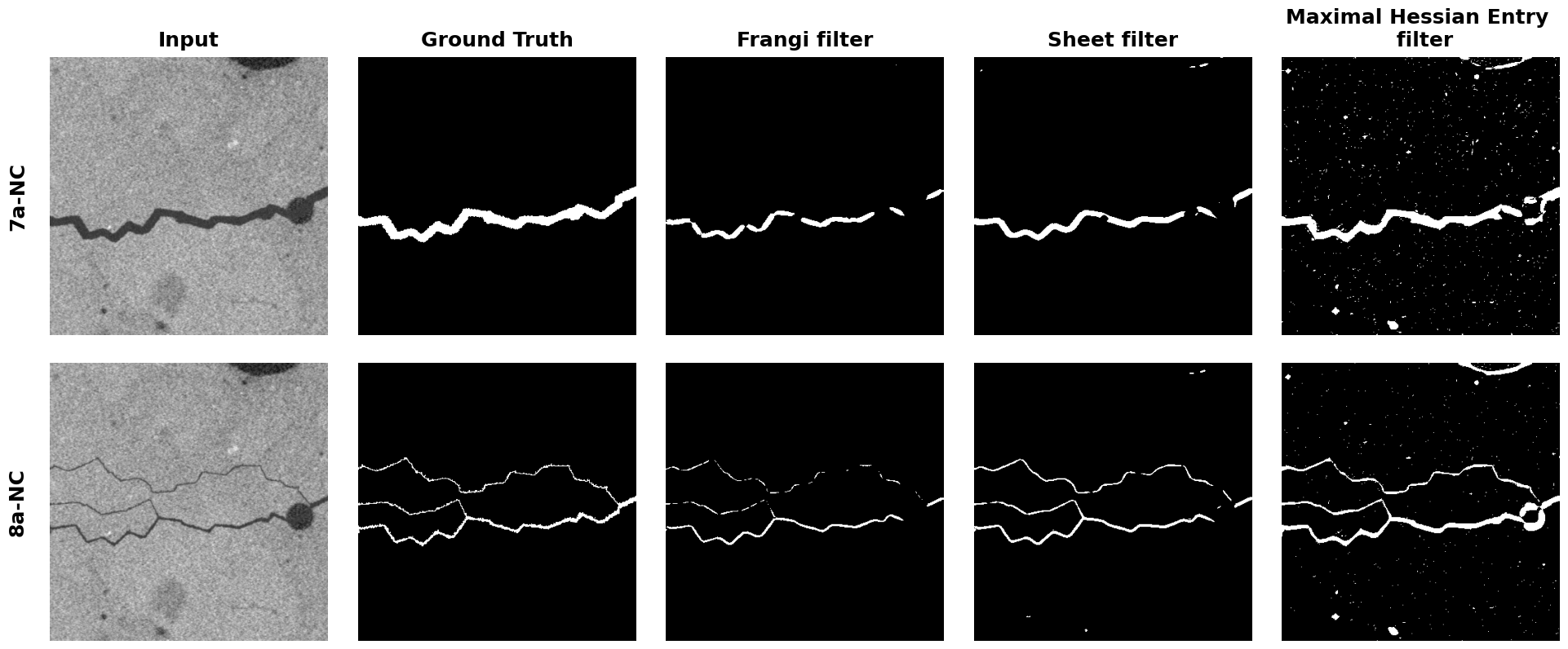}
	\caption{Results of crack segmentation on two semi-synthetic samples (7a-NC and 8a-NC) from the \textbf{VoroCrack3D} dataset \cite{JUNG2024110474}. 
		Each row corresponds to a test image (original and ground truth images), followed by the results of the Frangi filter, Sheet filter, and Maximal Hessian Entry filter. }
	\label{fig:segment}
\end{figure}

The results obtained by applying the Frangi filter, Sheet filter, and Maximal Hessian Entry filter to the selected semi-synthetic test images are shown in Figure \ref{fig:segment}. It is evident that all three approaches successfully capture the structural intricacies of cracks within the input images, which highlights the effectiveness of the Maximal Hessian Entry filter. However, due to the presence of pores with low gray values in the material, voxels belonging to the edges may be incorrectly identified as cracks. This discrepancy arises from the limited examination of the surrounding structure. Furthermore, by taking the maximum over all binary images, all noise is retained, resulting in a more noisy image compared to the ones from classical filters.

To provide a more comprehensive evaluation, the quality of these methods can be assessed through metrics such as Precision , Recall , F1-score ($F1$), and Intersection over Union (IoU), defined as follows:

\begin{equation}
	P=\frac{TP}{TP+FP}, \qquad
	R=\frac{TP}{TP+FN}, \qquad
	F_{1}=\frac{2P\cdot R}{P+R}, \qquad
	\mathrm{IoU}=\frac{TP}{TP+FP+FN}.
	\label{metrics}
\end{equation}

\begin{figure}[ht]
		\centering
		\includegraphics[width=\textwidth]{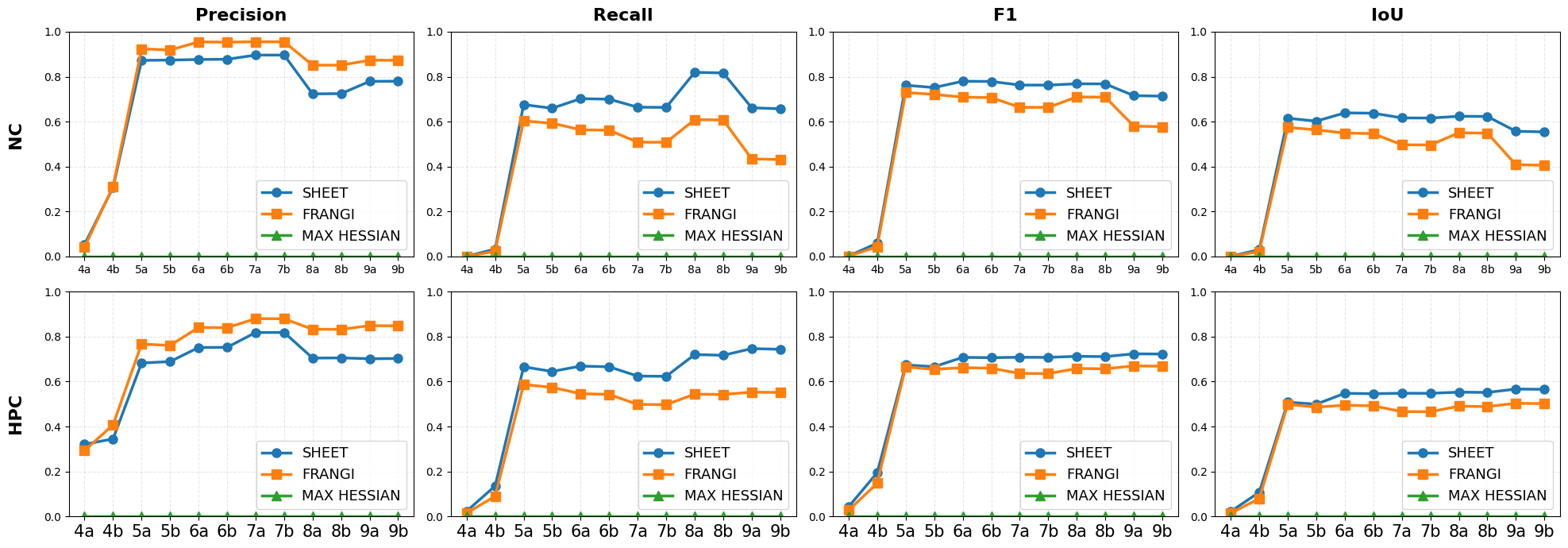}
		\caption{Performance of the Frangi, Sheet, and Maximal Hessian Entry filters evaluated using precision, recall, F1, and IoU. The top row shows results for normal concrete, and the bottom row shows results for high-performance concrete.}
		\label{fig:filtermenc}
	\end{figure}

Here, $TP$ (true positive) and $FP$ (false positive) represent the numbers of voxels correctly and falsely detected as belonging to a crack, respectively. Similarly, $TN$ (true negative) and $FN$ (false negative) denote the numbers of voxels correctly and falsely detected as material. 
Figures \ref{fig:filtermenc} illustrate the results for NC and HPC samples, respectively. 
Both the Frangi and Sheet filters demonstrate stable and consistent performance across all test samples with low noise levels and well-preserved crack patterns.
Between the two, the Sheet filter generally achieves slightly higher F1-score and IoU values, confirming its improved balance between precision and recall in the detection of thin to moderately wide cracks.
In contrast, it can be observed that the Maximal Hessian Entry filter achieves the highest recall values across most test cases, indicating its strong ability to capture crack voxels.
However, this comes at the cost of lower precision, F1 and IoU due to a higher number of false positives, particularly in regions affected by pores and noise.
This error arises from the simplicity of the Maximal Hessian Entry formulation, which relies primarily on local grayvalue gradients.
When combined with the maximum operator across multiple scales, the filter becomes highly sensitive to any local contrast changes, including those introduced by pores and other background fluctuations.
As a consequence, it captures almost all crack voxels, as indicated by its consistently high recall but this sensitivity frequently leads to false detections in non-crack regions.
	
\begin{figure}[ht]
		\centering
		\includegraphics[width=\textwidth]{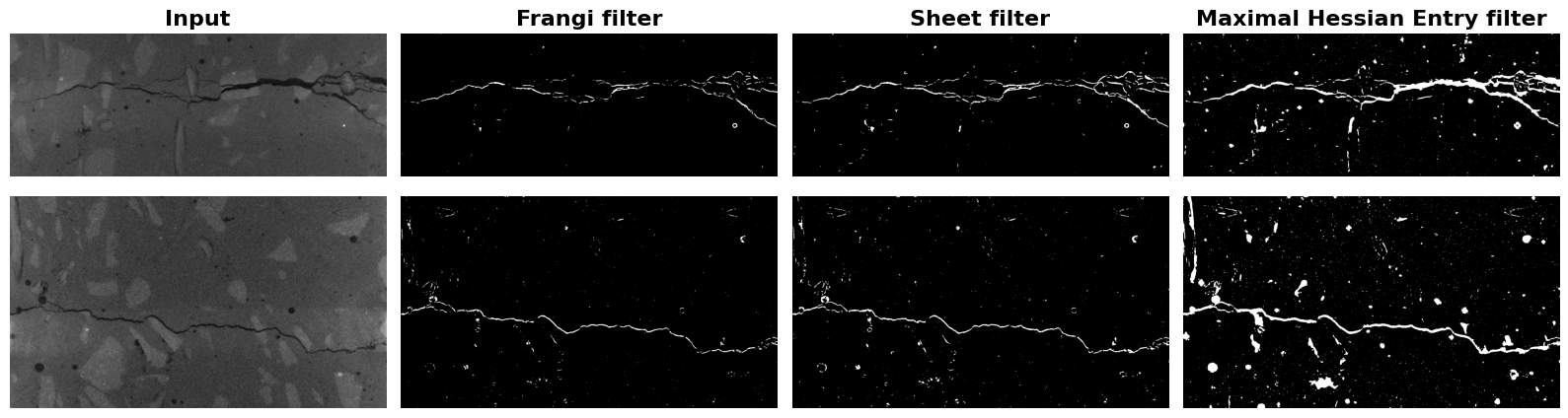}
		\caption{Application of the Frangi, Sheet, and Maximal Hessian Entry filters to real 3D CT images of concrete with cracks.
			Each row corresponds to a different input dimensions $850 \times 400 \times 1050$, $800 \times 650 \times 1150$, respectively.
			The first column shows 2D slices of the original images, followed by the segmentation results obtained using the three filters.
		}
		\label{fig:realfrangihessian}
	\end{figure}

Nevertheless, this limitation also shows the strength of the Maximal Hessian Entry filter.
Unlike other filters, which rely on a fixed $\sigma$ chosen in advance, the Maximal Hessian Entry approach integrates information across a set of scales, offering a potential advantage for real-world CT data, where the assumption of a constant crack width rarely holds. Here, we compare the Frangi, Sheet, and Maximal Hessian Entry filters applied to three volumetric datasets of different dimensions, see Figure \ref{fig:realfrangihessian}.
	
	The results in Figure \ref{fig:realfrangihessian} demonstrate the robustness of these filters when applied to real CT data.
	The Frangi and Sheet filters, both using $\sigma = 2.5$, can enhance well-defined cracks but often fail to detect parts of the cracks where the width or contrast changes.
	The Maximal Hessian Entry filter, on the other hand, combines information from several scales and therefore captures both thin and thick cracks more completely.
	However, this also makes the result noisier, as small pores and background variations are occasionally detected as cracks.
	Despite this, the method effectively preserves the connected crack patterns, which is important for materials where cracks differ in width and intensity.
	From a practical point of view, when the main objective is to preserve the overall crack structure, a simpler, inexpensive approach such as the Maximal Hessian Entry filter can be sufficient, especially for large 3D images where more complex multi-scale filters become prohibitively costly.

\subsection{Computational cost} \label{subsec:filtercost}

All three filters, Frangi, Sheet, and Maximal Hessian Entry share the same fundamental computational structure: the estimation of six second-order Gaussian derivatives to construct the Hessian matrix.  
If the Gaussian kernel has an effective width of $k = 4\sigma + 1$, the total cost of forming all six Hessian components is therefore
$
	F_{H} \approx 6\,(3 \cdot 2kN)
	= 36kN
	= 36(4\sigma+1)N .
$

For the Frangi and Sheet filters, an additional eigen-decomposition of the symmetric $3\times3$ Hessian is required at each voxel to obtain the principal curvatures. 
Theoretically, the arithmetic complexity of eigenvalue computation scales cubically with matrix dimension ($O(3^3)$) \cite{pmlr-v97-ghorbani19b}. Howerver, it depends on the diagonalization algorithm used \cite{orlowski2009efficient}, after which the filters require only a few algebraic operations and a voxelwise threshold to produce the binary response map. 
These steps introduce a modest additional cost of $c_{\text{post}} \approx 100$ FLOPs per voxel. 
In contrast, the Maximal Hessian Entry filter completely omits the eigen-decomposition step. 
After computing the six second-order derivatives across $S$ multiple scales, it performs a simple voxelwise maximum operation, followed by a mean-standard deviation-based threshold
These operations contribute only a few comparisons and additions per voxel, with a total cost denoted as $c_{\text{max}}$ (typically below 10 FLOPs). Therefore, the total cost becomes $F_{\text{MHE}}  = S(F_H + F_{\text{max}}) $.
In conclusion, the theoretical floating-point operation counts of the three filters can be summarized in Table \ref{tab:filter_flops}.

\begin{table}[h!]
	\centering
	\caption{Formulation of total floating-point operations per voxel for the three Hessian-based filters.}
	\label{tab:filter_flops}
	\begin{tabular}{lc}
		\hline
		\textbf{Filter} & \textbf{FLOPs} \\
		\hline
		Frangi & $F_{\text{Frangi}} = F_H + F_{\text{eig}} + F_{\text{post}} = 36(4\sigma + 1)N + (c_{\text{eig}} + c_{\text{post}})N$ \\
		Sheet & $F_{\text{Sheet}} = F_H + F_{\text{eig}} + F_{\text{post}} = 36(4\sigma + 1)N + (c_{\text{eig}} + c_{\text{post}})N$ \\
		Maximal Hessian Entry & $F_{\text{MHE}} = S(F_H + F_{\text{max}}) = S[36(4\sigma + 1)N + c_{\text{max}}N]$ \\
		\hline
	\end{tabular}
\end{table}

As shown in Table \ref{tab:filter_flops}, the Hessian computation term $F_H$ dominates the arithmetic complexity. However, despite the smaller nominal cost of the eigen-decomposition $F_{\text{eig}}$, in practice it often becomes the main computational bottleneck. The computation of Hessian eigenvalues is widely recognized as an inherently demanding task across disciplines ranging from large-scale optimization \cite{hare2023detecting} to biomedical imaging \cite{yang2014fast}, showing that the eigenvalue calculation often dominates the runtime of Hessian-based enhancement filters, since the operation must be repeated for every voxel and across multiple scales within multiscale frameworks \cite{yang2014fast,jerman2016enhancement}. Closed-form eigenvalue formulas for small Hessians are often numerically unstable, so accurate computation typically relies on slower iterative methods \cite{orlowski2009efficient}. This makes repeated eigenvalue evaluation a key bottleneck in Hessian-based approaches \cite{jerman2016enhancement}.
The Maximal Hessian Entry filter, by trading off precision for efficiency, provides a more scalable solution that preserves the essential crack topology at a fraction of the computational cost.

\subsection{Geometry of the binary image $L_\mathcal{S}$} \label{sec:geometry}
In this section, we propose a method to study the geometry of local structure elements, helping to distinguish cracks from noise such as air pores, stones, steel rods, etc. within concrete structures. By subdividing $W$ into small regions, we retain the essential crack geometry while limiting noise effects. The subimage size must be well calibrated to avoid losing geometric contrast. At an appropriate scale, cracks and noise exhibit reliably different geometry.

Let $L_\mathcal{S}: W \rightarrow \{0,1\}$ represent a binary image. 
First, we aim to construct a partition of cubes along the axes for the entire concrete structure. 
Let $W = N_1 \times N_2 \times N_3$, and let $g \in \mathbb{N}$ denote the edge length of each cube. 
The corresponding index set of cubes is
$
	\mathcal{W}_{g}
	= 
	\{1,\ldots, N_{1}/g\}
	\times
	\{1,\ldots, N_{2}/g\}
	\times
	\{1,\ldots, N_{3}/g\}.
$
The partition of $L_\mathcal{S}$ is defined as
$
	W = \bigcup_{q \in \mathcal{W}_{g}} W(q).
$
where all $W(q)$ are axis-aligned cubes of equal size. 
This results in a collection of cubes along the axes, denoted as $\{L_q : q \in \mathcal{W}_g\}$, where
$
	L_{q}(p)=\{L_{\mathcal{S}}(p), \ p\in W(q)\}.
$

Second, we compute various geometry statistics from the collection of subimages $\{L_q : q \in \mathcal{W}_g\}$. 
Intuitively, considering the continuity of cracks and their widths, we focus on subimages satisfying $|W(q)| \in [20^3,\, 30^3].$
The surface area density, defined as the ratio between boundary voxels and total voxels within a cube, captures the irregularity of local structures. Cracks, being thin and elongated, typically produce higher surface area density than compact objects such as pores or noise. 
The maximum connected region size quantifies the size of the largest contiguous component within each cube, providing a measure of local structural continuity. 
The foreground volume represents the total number of foreground voxels within a cube, measuring the amount of solid material present locally. 
Finally, the standard deviation of projected areas across 13 different 3D grid directions measures directional variability: crack regions typically show larger variations across directions due to their anisotropic shape, whereas nearly spherical pores or aggregates exhibit low variation.
Denoting $a_q$, $b_q$, $c_q$, and $d_q$ as the surface area density, maximum connected region size, foreground volume, and standard deviation of projected areas, respectively, we define
$
	T_{q} = (a_{q},\, b_{q},\, c_{q},\, d_{q}).
$
as a vector containing all statistics obtained from a subimage $L_q$. For more in-depth information on the definition of statistics, please refer to \cite{ohser20093d}. 
Based on these, we define the following four images:
\begin{equation}
	A_{g}=\{a_{q}\,:\, q\in\mathcal{W}_{g}\}, \
	B_{g}=\{b_{q}\,:\, q\in\mathcal{W}_{g}\}, \
	C_{g}=\{c_{q}\,:\, q\in\mathcal{W}_{g}\}, \
	D_{g}=\{d_{q}\,:\, q\in\mathcal{W}_{g}\}.
\end{equation}

It is crucial to observe that the obtained features have different ranges of values. Therefore, it is ideal to standardize each geometry statistic. At voxel $q\in\mathcal{W}_{g}$, $A^*_g$, $B^*_g$, $C^*_g$, and $D^*_g$ be defined as follows:
\begin{equation}
	A^{*}_{g}(q) = a_{q}\ \mathrm{sd}(A_{g})^{-1};\ B^{*}_{g}(q) = b_{q}\ \mathrm{sd}(B_{g})^{-1}; \ C^{*}_{g}(q) = c_{q}\ \mathrm{sd}(C_{g})^{-1}; \ D^{*}_{g}(q) = d_{q}\ \mathrm{sd}(C_{g})^{-1}.
\end{equation}

The derived statistics effectively discern disparities between anomaly regions and homogeneous regions, as the images $A^*_g$, $B^*_g$, $C^*_g$, and $D^*_g$ ought to replicate a similar crack pattern as the input CT images.
\begin{figure}[ht]
	\centering
	\includegraphics[width=\textwidth]{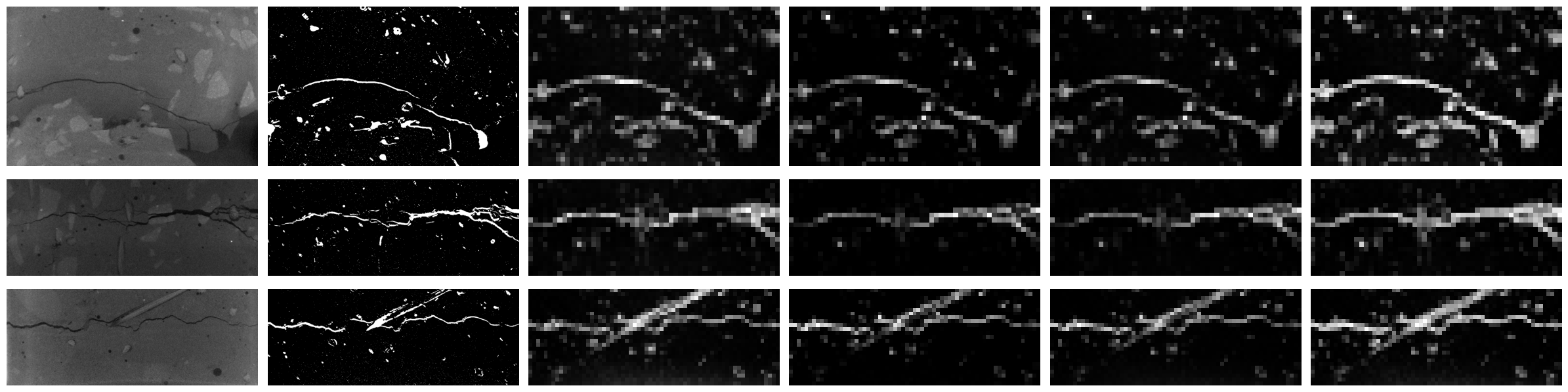}
\caption{
	Visualization of geometric statistics computed from different real input images at slice $z = 0$. 
	From left to right: the original image, the filtered binary image obtained by the Maximal Hessian Entry filter, and the four corresponding statistic maps of surface area density, maximum connected region size, foreground volume, and projection area standard deviation. 
}
	\label{fig:geometry}
\end{figure}
In Figure \ref{fig:geometry}, it is evident that the selected statistics serve as valuable features for identifying regions with cracks on several real-world CT datasets with varying dimensions, $840\times 400 \times 1040$, $800\times 700\times 1100$, and $920 \times 920 \times 1040$. Consequently, one can utilize these features for crack identification, offering an alternative to working with a large-scale input image. Moreover, one can handle all feature images  $A_g$, $B_g$, $C_g$, and $D_g$ as coordinate-wise realizations of a 4-variate random field $\mathcal{X} = \{ \xi_k \in \mathbb{R}^4, k \in  \mathcal{W} \}$. Within this framework, the classification of anomaly regions essentially involves studying the change-point problem of multivariate random fields.

\section{Crack detection as a change-point problem} \label{sec:anomaly}

In the book \cite{brodsky1993nonparametric}, the change-point problems are discussed under a general parametric setting using a CUSUM statistic test, which is widely accepted for anomaly detection, especially in time series under different dependence assumptions such as short-range or long-range dependence; cf.\ e.g.\ \cite{tartakovsky2014sequential,annika22}. Besides, it has been extended to 3D CT images of fibre materials to detect changes in the mean and entropy of local directional distributions \cite{Alonso-Ruiz_Spodarev_2017,AlonsoRuiz_Spodarev_2018}. 
For our problem, we apply a change-in-mean CUSUM statistic test to detect anomalies in the mean of a random field following the method described in \cite{Dresvyanskiy2020}, where tail bounds for the CUSUM statistic of stationary, $m$-dependent random fields were derived. The use of this test allows us to efficiently detect abrupt changes in the derived field of geometric crack descriptors while accounting for spatial dependencies in the data, and the resulting local statistics will later serve as input to the spatial multiple testing procedure.

\subsection{Multiple Hypotheses Testing} \label{subsec:MCP}
Let $\mathcal{X} = \{\xi_q = (\xi_q^{(1)}, \xi_q^{(2)}, \ldots, \xi_q^{(d)}), q \in \mathcal{W}\}, \mathcal{W} \subset \mathbb{Z}^n$ denote a stationary, centered, vector-valued random field, $\mathbb{P}\{\xi_q \in \mathbb{R}^d\} = 1$. We introduce the finite parameter space $\Theta$, where for each $\theta \in \Theta$, we define $J_\theta \subset \mathcal{W}$ as the corresponding scanning window within $\mathcal{W}$. The selection of $J_\theta$ is crucial, and it should avoid being excessively small or large, ensuring meaningful analysis. The change-point hypotheses for any $J_\theta, \theta \in \Theta$, are formulated as follows:

$H_0(\theta): \mathbb{E}\xi_k = \mu \in \mathbb{R}^d$ for all $k \in W$, i.e. there is no change in mean, versus

$H_1(\theta)$: there exists a vector $h \in \mathbb{R}^d, h \neq 0$ such that $ \mathbb{E}\xi_k = \mu + h, k \in J_\theta$ and $ \mathbb{E}\xi_q = \mu , q \in J^c_\theta$.

To test $H_0(\theta)$ versus $H_1(\theta)$, for $p \geq 1$, employ the change-in-mean CUSUM statistic:
\begin{equation}
	T(\theta)
	= 
	\max_{i=1,\ldots,d}
	\left(
	\frac{1}{|J_{\theta}|}\sum_{k\in J_{\theta}} \xi_{k}^{(i)}
	-
	\frac{1}{|J^{c}_{\theta}|}\sum_{k\in J^{c}_{\theta}} \xi_{k}^{(i)}
	\right).
	\label{eq:ttheta}
\end{equation}

The null hypothesis $H_0(\theta)$ is rejected when the test statistic $T(\theta)$ exceeds a pre-specified critical value denoted as $y(\theta, \alpha)$, with $\alpha$ representing the predetermined significance level. Alternatively, if the $p$-value $p(\theta)$ associated with the statistic $T(\theta)$ is available, the null hypothesis $H_0(\theta)$ is rejected when $p(\theta) \leq c(\theta,\alpha)$.
 In order to find the $p$-value, knowing the null distribution, i.e. the distribution of $T(\theta)$ given $H_0$ holds is required.
However, in the absence of knowledge about the distribution, determining the $p$-values becomes challenging. To address this issue in practical applications, the empirical null distribution for $T(\theta)|H_0$ can be used.

To identify regions containing cracks, one needs to test all null hypotheses $H_0(\theta), \theta \in \Theta$, simultaneously. Analogous to single hypothesis testing, where controlling Type-I error and Type-II error  at a certain level $\alpha$ are crucial, in the context of multiple testing, there are two types of errors known as family-wise error (FWER) and false discovery rate (FDR). Assuming $|\Theta| = M$ is the total number of hypotheses, let $FP$ or $TP$ represent the total number of falsely or correctly rejected null hypotheses. FWER and FDR are defined as follows:
\begin{equation}
	FWER = \mathbb{P}\{FP > 0\} \text{ and } FDR = \mathbb{E}\left[\frac{FP}{FP+TP}\right].
	\label{eq:errors}
\end{equation}

In other words, FWER is the probability of committing one or more false discoveries within a family of tests while FDR is the expected value of the proportion of false positives among the rejected null hypotheses.
In crack detection scenarios with a large number of hypotheses ($M$ is very large), the objective is often to maximize discoveries. Early approaches, such as the Šidák correction \cite{sidak67}, aimed to rigorously control the Family-Wise Error Rate (FWER) to prevent inflation of the probability of making at least one false discovery. Subsequent methods, introduced by \cite{hochberg1988, holm79}, included step-down and step-up procedures that dynamically adapt based on observed p-values, striking a balance between error control and sensitivity. In large-scale hypothesis testing in general, and in crack detection in particular, one can relax the rejection rule, allowing for more false positives and increased power. As an alternative, controlling FDR allows for a higher level of false $H_0$ rejections while enhancing test power \cite{Benjamini1995,sarkar2022local,cao2022optimal}. The well-known Benjamini-Hochberg procedure \cite{Benjamini1995}, a step-up approach, effectively controls FDR at a pre-specified level $\alpha$ under the assumption of independence among hypotheses. However, LAWS procedure \cite{cai2021} offers an alternative to settings where independence or weak dependence among hypotheses does not hold, allowing modifies p-values adaptively by incorporating local sparsity information, opening for more precise adjustments in spatially or temporally dependent data. This is a flexible, data-driven approach that deals with independent and dependent cases effectively, offering an alternative to methods like AdaPT \cite{lei2018} and SABHA \cite{lit2019}, which require substantial prior knowledge or complex modeling. For each hypothesis $\theta \in \Theta$, the locally adjusted $p$-value following the construction of \cite{cai2021} is defined as $
	p^w(\theta) = \min\left(1, \frac{p(\theta)}{w(\theta)}\right),
$
where $w(\theta)$ is a weight determined by the local sparsity, given by:
$
	w(\theta) = \frac{1 - \hat{\pi}(\theta)}{\hat{\pi}(\theta)},
	\label{eq:weights}
$
and $\hat{\pi}(\theta)$ is an estimate of the proportion of true null hypotheses in the local neighborhood of $\theta$. The local sparsity $\hat{\pi}(\theta)$ is computed using a kernel-based approach:
\begin{equation}
	\hat{\pi}(\theta) = 
	\frac{\sum_{\theta' \in T_\tau} K_h(\theta - \theta')}
	{(1 - \tau)\sum_{\theta' \in \Theta} K_h(\theta - \theta')}.
	\label{eq:localpi}
	\end{equation}

where $T_\tau = \{\theta' \in \Theta: p(\theta') > \tau\}$ is the set of hypotheses which is likely to be null, $ 
K_h(t)$ is a kernel function with bandwidth $h$, and $\tau$ is a threshold parameter. In this paper, we use the standard Gaussian kernel 

\begin{equation}
	K_h(t) = \frac{1}{\sqrt{2\pi}h} \exp\left(-\frac{t^2}{2h^2}\right).
	\label{eq:kernelh}
\end{equation}

By weighting hypotheses adaptively based on the local sparsity, LAWS effectively accounts for dependencies in the data, making it particularly well-suited for spatially dependent settings such as crack detection. The rejection rule now is defined as follows:
Reject $H_0(\theta)$ if $p^w(\theta) \leq \alpha$.

We apply this adaptive weighting procedure to both semi-synthetic and real concrete CT images to assess its practical performance. In the following numerical section, we demonstrate how it efficiently localizes potential crack regions within the dataset. As emphasized earlier, the goal is not to perform full segmentation but to identify regions of interest where cracks are likely to occur. Therefore, the numerical experiments focus on evaluating the accuracy and robustness of the pre-localization step, rather than comparing against other deep learning-based segmentation methods. 

\section{Experimental evaluation} \label{subsec:numeric}
\subsection{Data acquisition}\label{sec:data-acquisition}
The experimental evaluation is based on two complementary types of three-dimensional CT data: semi-synthetic volumes and real concrete scans.

The semi-synthetic images originate from the VoroCrack3D dataset \cite{JUNG2024110474}.
In this study, we selected a total of 48 volumes corresponding to cases 4a-4d through 9a-9d from both the NC and HPC subsets. 
These images cover a range of crack widths and configurations and therefore serve as a control benchmark for evaluating detection performance under varying structural and noise conditions.

The real CT data consist of eight concrete samples scanned using the large-scale CT system \cite{salamon2025gulliver} at University of Kaiserslautern.
This high-energy gantry-type scanner is designed for full-scale construction components and can acquire voxel sizes from about 50\,\textmu m to 300\,\textmu m, covering the resolution range required to detect pores and microcracks on the order of 0.1\,mm.
The real scans used in this study were obtained under typical operating conditions of the Gulliver system and contain natural concrete features such as porosity, natural material texture and small internal cracks, providing a realistic setting for evaluating the proposed statistical methodology.

\subsection{Empirical null modelling}

In the statistical framework used in this study, the null distribution of the test statistic $T(\theta)$ in (\ref{eq:ttheta}) is not available in closed form, so it must be estimated directly from the data. 
Before applying any multiple testing procedure, one must ensure that the $p$-value calculation reflects the characteristics of the material; otherwise, an incorrect null estimate can cause substantial errors in the testing step and lead to misleading detection results.
To address this, we first analyze only one homogeneous $400^3$ CT image without cracks and use it to compute the distribution of $T(\theta)$ under the null hypothesis of no anomalies. 
Relying on a single clean scan is also practical, since obtaining multiple crack-free concrete scann from the \cite{salamon2025gulliver} system is typically impractical to arrange and comes with high resource requirements.

\subsection{Testing procedure} \label{null-estimation}

For the computation of the geometry statistics, the input image of size 
$N_1 \times N_2 \times N_3$ is subdivided into cubes of edge length $g = 20$. Using cubes that are much smaller or much larger would make the geometric features 
unreliable and could lead to misclassification. This yields a regular grid of 
$
\mathcal{W}_g = 
\{1,\ldots, N_1/g\}
\times
\{1,\ldots, N_2/g\}
\times
\{1,\ldots, N_3/g\},
$
where each $q \in \mathcal{W}_g$ indexes one cube $W(q)$ in the partition.

The scanning window size is set to $u=3$to ensure anomaly proportions remain within 5–50\%, prevents cracks from overwhelming the window or becoming negligible when they appear only along an edge. The choice $u=3$ therefore provides a balanced window size in which the anomaly proportion remains in a meaningful range for reliable testing.
Since the size of the scanning window is fixed at $u=3$, and the corresponding family of windows is defined as 
$
\Theta = \{\theta = (a,b,c,u) : a,b,c \in \mathbb{Z},\;
J_\theta = [a,a+u] \times [b,b+u] \times [c,c+u] \subset \mathcal{W}_g \}.
$
The empirical null distribution $\hat{F}$ is obtained by computing the statistics $T(\theta)$ for all scanning windows $\theta \in \Theta$ within the homogeneous $400^3$ image using the same settings, and then taking the empirical cumulative distribution function of these values. 
For a test image, the $p$-value of the test statistic $T(\theta)$ at $\theta \in \Theta$ is given by $p(\theta) = 1 - \hat{F}(T(\theta))$.

Because the scanning windows overlap, the decision for each region of the image depends on several hypothesis tests. 
Let $W = \bigcup_{q \in \mathcal{W}_g} W(q)$ denote the spatial partition of the original image into cubes and the scanning windows are applied not to the voxels but to the index grid $\mathcal{W}_g$, where each position corresponds to one geometric statistics cube. 
For each cube $W(q)$, let $\mathcal{I}_q$ be the set of scanning windows $J_\theta$ whose index sets contain the index $q$.
For example, when using a window size of $u=3$, the window spans three cube indices along each axis in $\mathcal{W}_g$. 
Thus, a single cube $W(q)$ may be included in up to $27$ different scanning windows.
 
After performing multiple testing at level $\alpha$, we obtain the set of local decisions 
$
\mathcal{R}_{\alpha,q} = \{ R_\alpha(\theta) : J_\theta \in \mathcal{I}_q \},
$
where $R_\alpha(\theta) = 1$ indicates that the null hypothesis $H_0(\theta)$ is rejected (evidence of an anomaly), and $R_\alpha(\theta) = -1$ indicates that $H_0(\theta)$ is accepted (no anomaly detected). A cube $W(q)$ is classified as anomalous if $\sum_{x \in \mathcal{R}_{\alpha,q}} x \ge 0.$

\subsection{Semi-synthetic images} \label{subsec:nummericsyn}

\begin{figure}[H]
	\centering
	\includegraphics[width=0.8\textwidth]{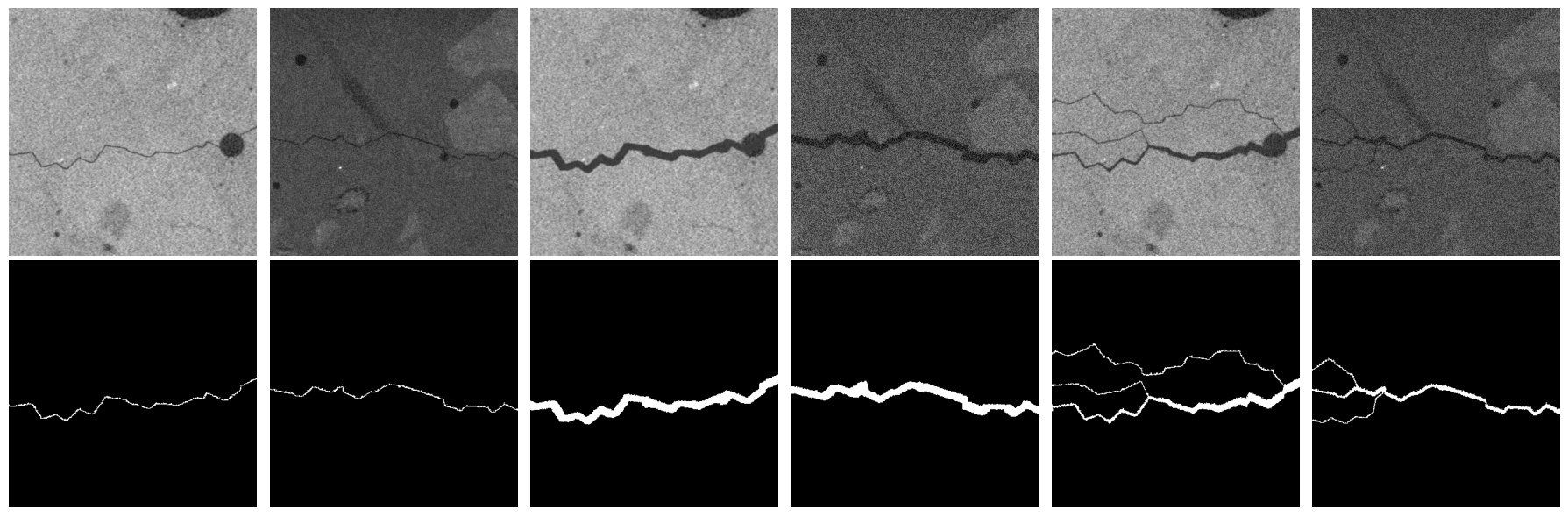}
	\caption{
		Representative 2D slices from several semi-synthetic images in the VoroCrack3D dataset.
		Top row: input CT slices with different crack widths, orientations, and noise levels. 
		Bottom row: corresponding ground-truth.
	}
	\label{fig:synexamples}
\end{figure}

In this study, we use a total of 48 volumes corresponding to cases 4a-4d through 9a-9d 
from both the NC and HPC subsets. 
These images exhibit a wide range of crack widths, orientations, and noise levels, 
providing a controlled and diverse environment for assessing the behaviour of the proposed 
procedure, see Figure \ref{fig:synexamples}.
The testing framework introduced in Section \ref{null-estimation} is applied uniformly to all semi-synthetic volumes. 
For each image, the geometric statistics are computed on cubes of edge length $g = 20$, and the resulting fields are scanned using windows of size $u = 3$. 
The empirical null distribution $\hat{F}$, estimated from the homogeneous $400^3$ image, is then used to compute the CUSUM statistics $T(\theta)$ into $p$-values. 
Final anomaly decisions are obtained by aggregating the outcomes of all overlapping scanning windows.

To investigate the effect of the tuning parameters on detection performance, the method 
was evaluated over a grid of values for the 
global FDR level $\alpha$ and the threshold $\tau$.  
Let $\alpha \in [0.05,\,0.30]$ and $\tau \in [0.05,\,0.70]$, yielding 
$11 \times 27 = 297$ parameter combinations.    
These parameters enter the weighting scheme through the local sparsity estimator 
\eqref{eq:localpi}, together with the kernel bandwidth $h$ in \eqref{eq:kernelh}.  
The bandwidth was fixed at $h = 1$ so that the standard Gaussian kernel is used and the sparsity estimator remains strictly local. If $h$ were chosen larger, the kernel would place substantial weight on a wide spatial neighborhood. Because cracks typically form planar structures, they occupy only a small fraction of this larger neighborhood, causing the sparsity estimate $\hat{\pi}(\theta)$ to become too large, increasing the modified $p$-values and decreasing sensitivity along the crack. 

\begin{figure}[ht]
	\centering
	\includegraphics[width=\textwidth]{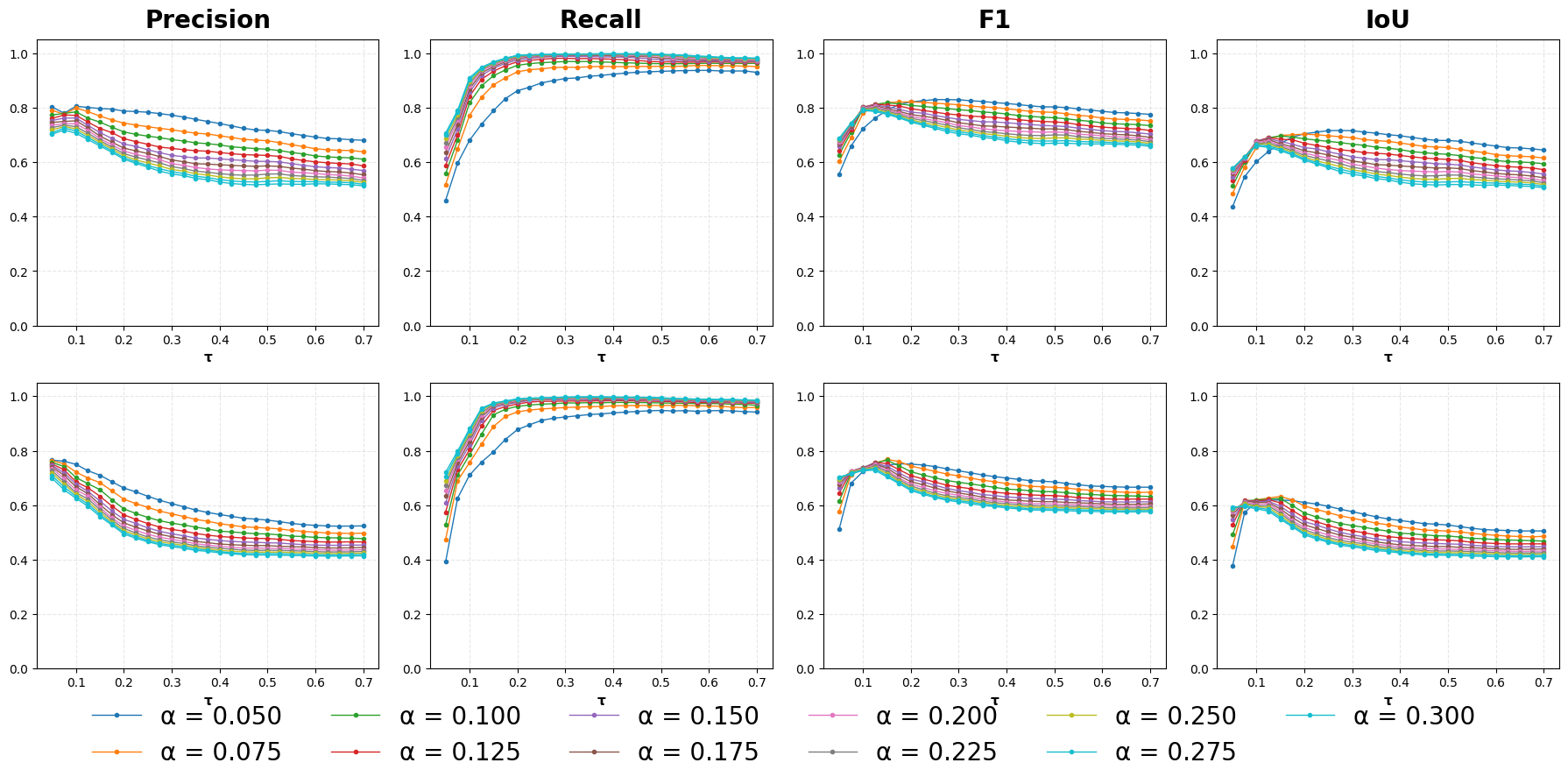}
	\caption{Mean precision, recall, F1, and IoU as functions of the threshold parameter $\tau$ for all values of $\alpha$ in the grid. The top row corresponds to the NC dataset and the bottom row corresponds to the HPC dataset. .}
	\label{fig:testmetric}
\end{figure}

For each volume and each pair $(\alpha,\tau)$ in the grid, the resulting binary decision field was compared with the corresponding ground truth image, and the standard metrics precision, recall, F1 and IoU were computed, following the definitions in (\ref{metrics}). The metrics were then averaged separately over the NC and the HPC sets. The resulting performance profiles over the $(\alpha,\tau)$ grid are shown in Figure \ref{fig:testmetric}, exhibiting a consistent pattern across both the NC and HPC datasets. The F1-score and IoU curves rise sharply for small values of $\tau$, achieve their highest values in the narrow region around $\tau = 0.1$, and subsequently fall due to the gain in recall becoming too small compared with the reduction in precision.

\begin{figure}[ht]
	\centering
	\includegraphics[width=\textwidth]{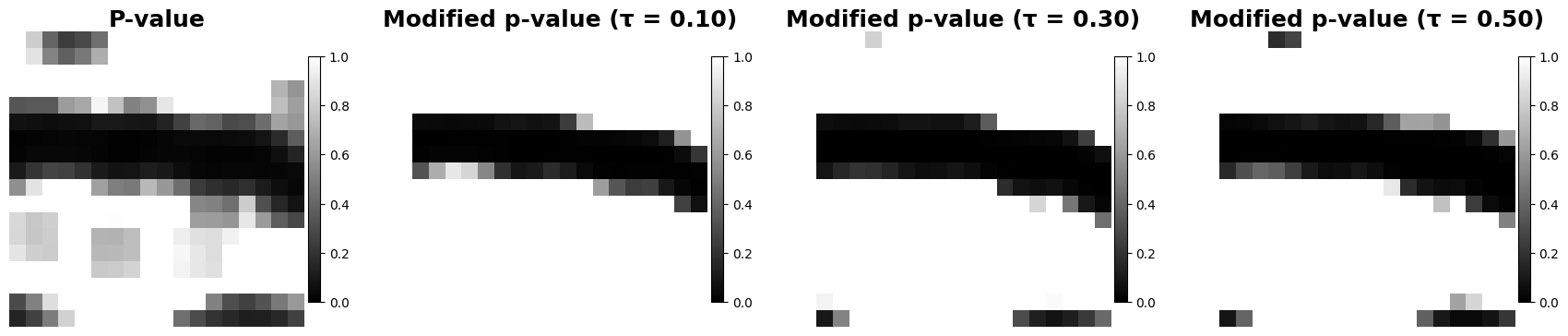}
\caption{Visualisation the raw $p$-values, and the modified $p$-values for several choices of the parameter $\tau$.}

	\label{fig:inspecttau}
\end{figure}

The influence of the parameter $\tau$ follows immediately from the local sparsity estimator in \eqref{eq:localpi}. As $\tau$ increases, the set $T_\tau = \{\theta' : p(\theta') > \tau\}$ shrinks, which reduces the value of $\hat{\pi}(\theta)$ and therefore decreases the modified $p$-values in general. An example of this effect is shown in Figure \ref{fig:inspecttau}.

With $\tau \approx 0.1$, the method attains its best overall performance: F1-score reaches approximately $0.78$-$0.80$ and the IoU about $0.65$-$0.70$ on the NC dataset, with slightly lower maxima on the HPC data, while recall is already high for nearly all choices of $\alpha$, typically around $0.85$-$0.93$ across both datasets, indicating that almost all crack regions are detected. Precision at $\tau = 0.10$ lies between $0.70$ and $0.78$ for NC and between $0.65$ and $0.75$ for HPC when $\alpha$ is in the range $0.10$-$0.20$, showing that this choice offers a balanced trade-off between sensitivity and false discoveries. 
Although $\tau = 0.1$ may seem low for a $p$-value threshold, it is appropriate for the geometry of the scanning windows. Windows centered on anomalous cubes produce very strong statistics and thus $p$-values close to zero. When the window moves slightly away from the crack, it still overlaps a part of the crack region, so the mean statistic remains above the null level and the resulting $p$-values are only moderately small. These partially overlapping windows would reduce the ability to distinguish crack from background if the threshold were set higher. Choosing $\tau = 0.1$ cleanly separates such weakly overlapping windows from genuinely null ones, which produce $p$-values well above this level. This explains why $\tau = 0.1$ consistently yields the best performance: it captures the full extent of the crack trajectory while avoiding false positives in regions where the window barely or no longer overlaps the crack.

 Using the definition of the false discovery rate in \eqref{eq:errors}, the empirical 
 false discovery rate $\mathrm{FDR} =  FP /(TP+FP)$ is exactly equal to $1 - Precision$. At $\tau = 0.10$, this empirical 
 FDR is approximately $22\%$-$30\%$ for NC and $25\%$-$35\%$ for HPC. For $\alpha$ in the range 
 $0.10$-$0.20$, the corresponding precisions lie between $0.75$ and $0.78$ on both 
 datasets. In practice, values of $\alpha$ in this interval are the most realistic, since the underlying dependence structure of the test statistics is not fully characterized, and choosing $\alpha = 0.05$ or $0.10$ typically produces an overly conservative procedure, leading to a significant drop in recall. Therefore, taking $\alpha$ around $0.20$ ensures that the FDR remains controlled and that the test retains substantial power under the observed dependence structure.
 
\begin{figure}[ht]
 	\centering
 	\includegraphics[width=\textwidth]{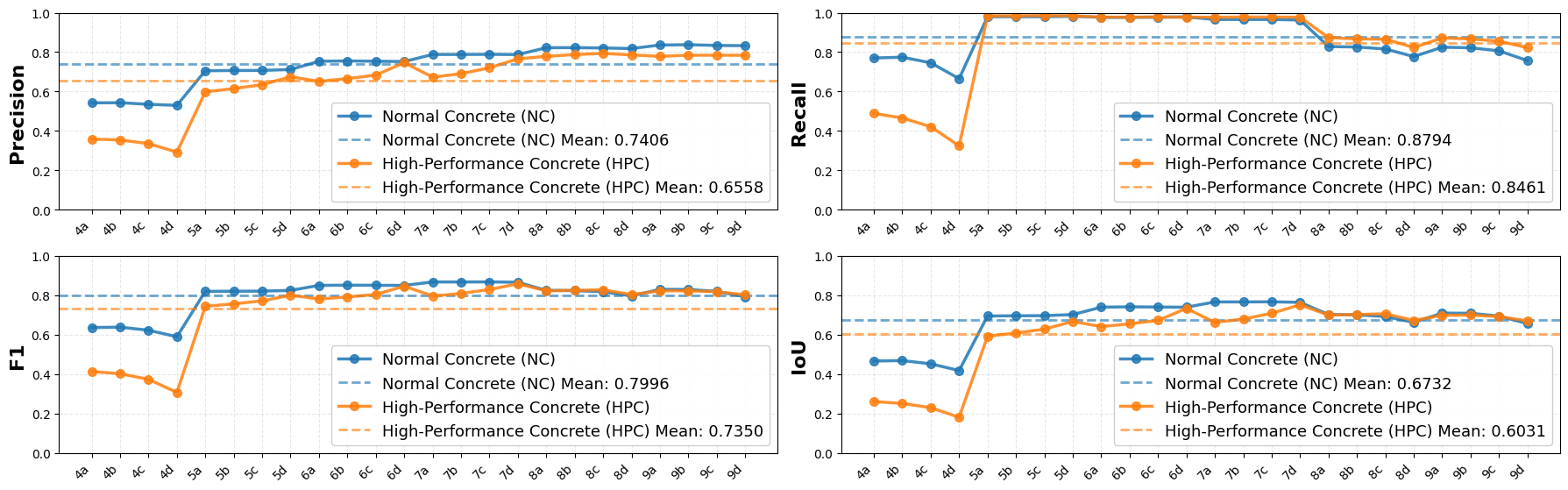}
 	\caption{Per-image performance for all 48 semi-synthetic volumes under the
 		fixed parameter choice $\alpha = 0.20$, $\tau = 0.10$, and $h = 1$. Precision,
 		recall, F1-score, and IoU are shown for each NC (blue) and HPC (orange)
 		image, together with their dataset-wise means (dashed lines).}
 	\label{fig:syntestall}
 \end{figure}

 To complement this qualitative example, Figure \ref{fig:syntestall} shows the
 per-image performance across all 48 semi-synthetic volumes under the same
 parameter choice $(\alpha = 0.20,\;\tau = 0.10,\;h = 1)$. The four metrics
 (precision, recall, F1, and IoU) are displayed for each image separately
 and compared between the NC and HPC datasets. The behaviour is stable across the
 entire collection, with NC images achieving slightly higher values on average
 and HPC images following the same trend with mildly reduced precision and IoU.

\begin{figure}[H]
	\centering
	\includegraphics[width=0.97\textwidth]{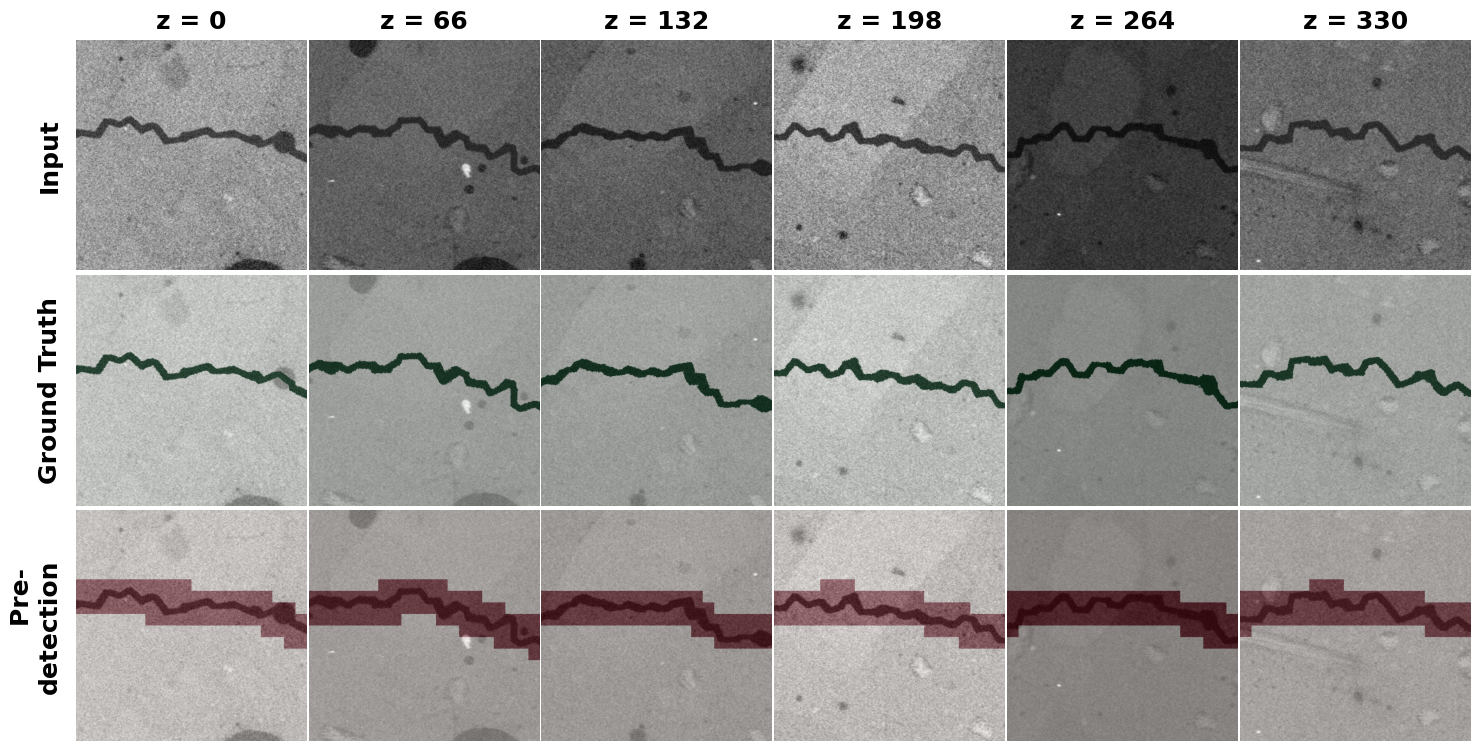}
	\caption{Example of crack localization with parameters fixed at
		$\alpha = 0.20$, $\tau = 0.10$, and $h = 1$. The top row shows six input slices at
		different depths; the middle row displays the corresponding ground-truth; and the bottom row presents the predicted detection maps.}
	\label{fig:syntest}
\end{figure}

With the parameters fixed at $\alpha = 0.20$, $\tau = 0.10$, and $h = 1$, the resulting
binary crack detections remain closely aligned with the
ground-truth. A representative example displayed in
Figure \ref{fig:syntest} corresponds to the 7b-NC image (width $w=7$), showing that the method successfully localizes the crack across the image, with only a small increase in the rejection volume, reflecting the moderate false discovery rate quantified earlier. Here we use the green marks as the ground truth for semi-synthetic images. For all visualizations in the remainder of the paper, the red shaded region represents the subset of the regions identified as anomalies by our pre‐localization method.

\begin{figure}[H]
	\centering
	\includegraphics[width=0.97\textwidth]{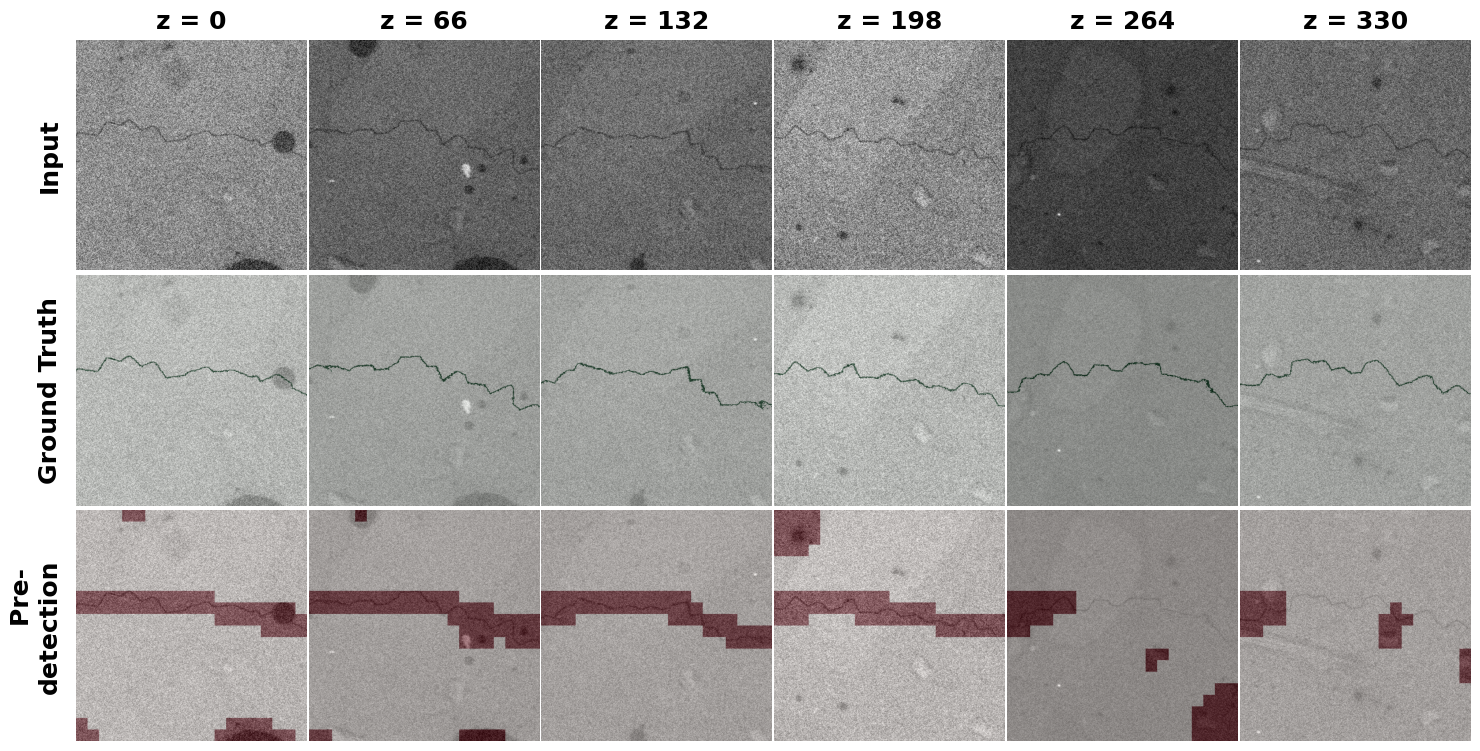}
	\caption{Example of a challenging case from the NC dataset (image 4d).}
	\label{fig:4cNC}
\end{figure}

A noticeably different behaviour occurs for the image 4d-NC, where the 
crack is extremely thin, embedded in strong 
noise, which explains its markedly lower 
scores in Figure \ref{fig:syntestall}. In this situation the geometric statistics fail to capture the crack 
regions, causing the CUSUM values along the anomalous regions to be indistinguishable from homogeneous areas. This leads to incorrect estimation of 
$p$-values. After the weighting step, 
the modified $p$-values no longer focus around the crack, and the rejection set 
becomes unstable, fragmented, or absent. This example demonstrates how thin cracks geometry combined with a low 
signal-to-noise ratio can cause the method to fail to perform a reliable detection, see Figure \ref{fig:4cNC}.

Overall, the method consistently recovers the underlying cracks across most semi-synthetic cases, including thin, thick, and multiscale configurations, and remains stable under varying noise levels. These results demonstrate that the testing framework behaves predictably under controlled conditions before being applied to real CT datasets.

\subsection{Real CT images of concrete}\label{subsec:nummericreal}
In the second part of the experimental study, the methodology is applied to the real CT images acquired with the Gulliver CT scanner described in Section \ref{sec:data-acquisition}. These scans contain natural concrete structure, pores and texture, and therefore serve as a realistic examination of the method under the full material complexity of concrete.

\begin{figure}[ht]
	\centering
	\hspace*{-0.6cm}
	\includegraphics[width=1.0\textwidth]{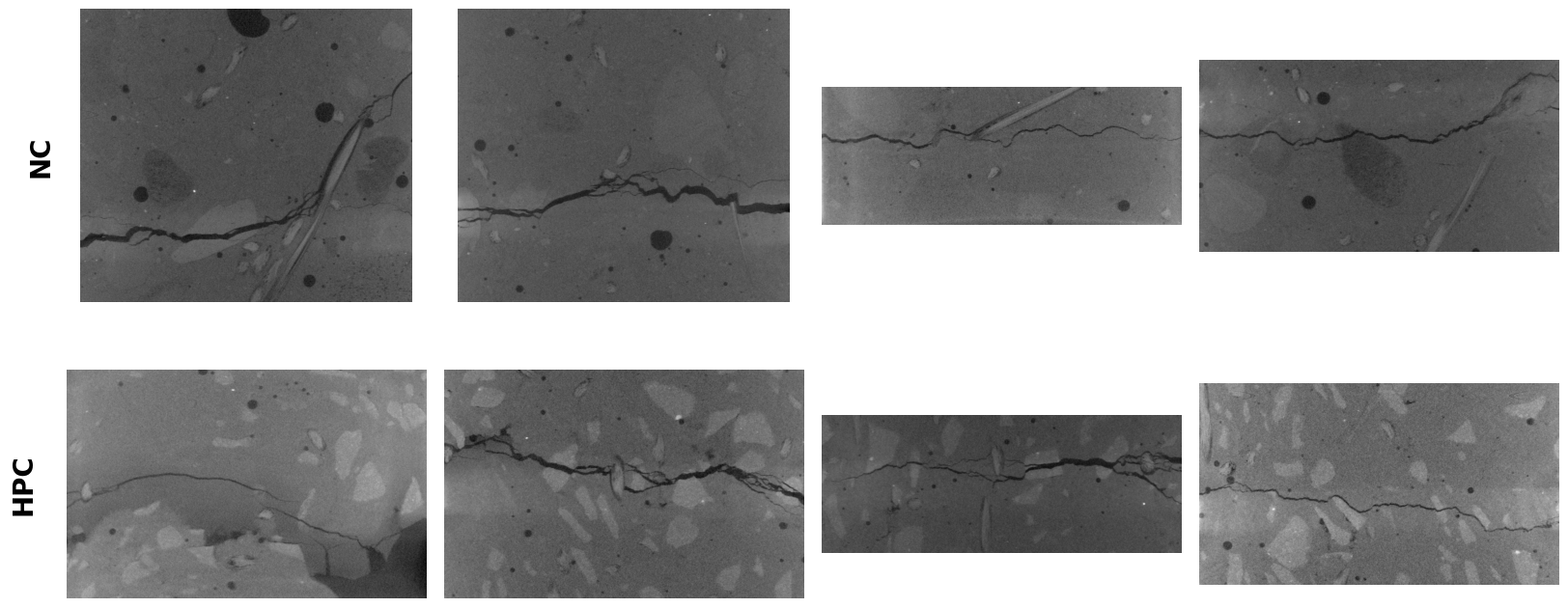}
\caption{Representative slices of Normal Concrete (NC-1, NC-2, NC-3, NC-4; top row) 
	and High-Performance Concrete (HPC-1, HPC-2, HPC-3, HPC-4; bottom row), arranged 
	from left to right.}

	\label{fig:nchpc}
\end{figure}

The real-data evaluation uses eight CT volumes, four from normal concrete and four from high-performance concrete. Their voxel dimensions range from 
approximately $640 \times 400 \times 1040$ to $1000 \times 920 \times 1300$, covering 
the typical size spectrum of large concrete scans produced by the Gulliver system. 
To simplify notation, we refer to the scans as  NC-1,…,NC-4 and HPC-1,…,HPC-4, see Figure \ref{fig:nchpc}.

\begin{figure}[ht]
 	\centering
 	\includegraphics[width=0.60\textwidth]{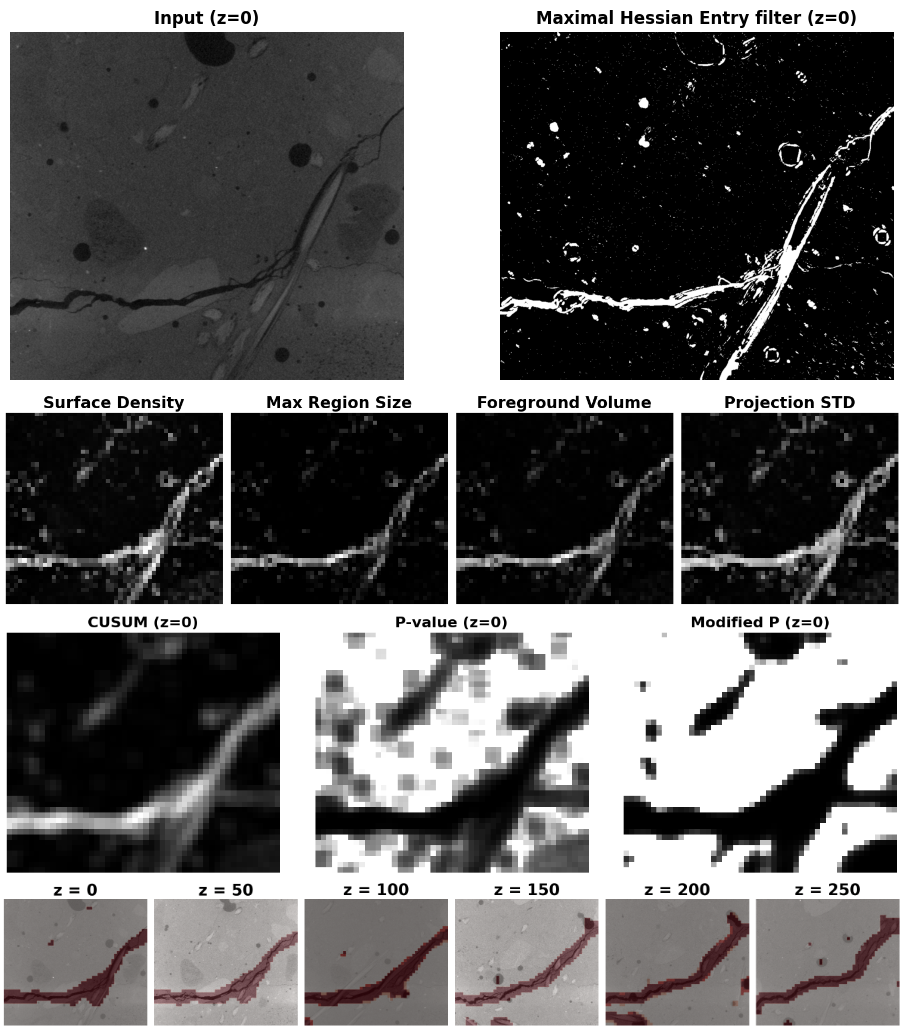}
 	\caption{
 		Illustration of the full crack localization pipeline on the real 3D CT 
 		volume NC-1. Top row: input slice and maximal Hessian entry filter. 
 		Second row: geometric statistics (surface density, maximal region size, 
 		foreground volume, and projection standard deviation). 
 		Third row: CUSUM statistic, raw $p$-values, and modified $p$-values for 
 		$\tau = 0.10$. 
 		Bottom row: resulting binary anomaly maps over several depths. 
 	}
 	\label{fig:methodnc1}
 \end{figure}

 The crack localization procedure for the real CT data uses the previously selected parameters $g = 20$, $\alpha = 0.20$, $\tau = 0.10$, and $h = 1$. To illustrate how the procedure performed on real CT data, 
 Figure \ref{fig:methodnc1} shows the full workflow of the procedure applied to the NC-1 image, including all intermediate quantities and the final crack localization. For the NC-1 scan, the pipeline produces well-localized regions containing cracks.  
 The CUSUM statistic forms a clear band over the anomalous areas, and the weighting step suppresses homogeneous regions, leading the modified $p$-values to to be low on the cracked ares.  
 The resulting rejection set matches the visible crack structure throughout the depth of
 the sample and remains stable even where the crack becomes thin.  
 This demonstrates that the procedure maintains sensitivity on real CT data despite the
 presence of material texture and noise.
 
 A direct quantitative evaluation is not possible for the real CT data, since
 voxel-level ground truth is not available. Cracks in these volumes are natural
 microstructural features, and producing manual three-dimensional segmentation is
 not feasible. As a reference, we use crack segmentations produced by a three-dimensional U-Net 
 following the architecture of \cite{nowacka2024deep}, trained on 
 the VoroCrack3D dataset by our collaborators at University of Kaiserslautern. The network is a 
 standard 3D U-Net with three downsampling levels and an initial feature width 
 of 16, comprising approximately 1.4 million trainable parameters. These 
 segmentations are not voxel-accurate ground truth, but they provide a stable 
 and structurally reliable benchmark against which the statistical method can be compared. However, as we mentioned earlier, the two methods serve fundamentally different purposes: the neural network performs full crack segmentation, while the proposed procedure provides a  localization of anomalous regions which might be used for training, pre-detection or independent cross-checking. The comparison therefore examines whether the statistical test highlights regions that the U-Net also mark some voxels inside as cracks, and whether there exist areas where the U-Net ignores a crack within but the statistical procedure shows that there are anomalies, or where the U\textsc{-}Net produces a segmentation but the statistical procedure classifies the same region as homogeneous.
 
 A direct quantitative evaluation is not possible for the real CT data, since
 voxel-level ground truth is not available. Cracks in these volumes are natural
 microstructural features, and producing manual three-dimensional segmentation is
 not feasible. As a reference, we use crack segmentations produced by a three-dimensional U-Net 
 following the architecture of \cite{nowacka2024deep}, trained on 
 the VoroCrack3D dataset by our collaborators at University of Kaiserslautern. The network is a 
 standard 3D U-Net with three downsampling levels and an initial feature width 
 of 16, comprising approximately 1.4 million trainable parameters. These 
 segmentations are not voxel-accurate ground truth, but they provide a stable 
 and structurally reliable benchmark against which the statistical method can be compared. However, as we mentioned earlier, the two methods serve fundamentally different purposes: the neural network performs full crack segmentation, while the proposed procedure provides a  localization of anomalous regions which might be used for pre-detection or independent cross-checking. The comparison therefore examines whether the statistical test highlights regions that the U-Net also mark some voxels inside as cracks, and whether there exist areas where the U-Net ignores a crack within but the statistical procedure shows that there are anomalies, or where the U\textsc{-}Net produces a segmentation but the statistical procedure classifies the same region as homogeneous.
 
\begin{figure}[H]
 	\centering
 	\includegraphics[width=0.8\textwidth]{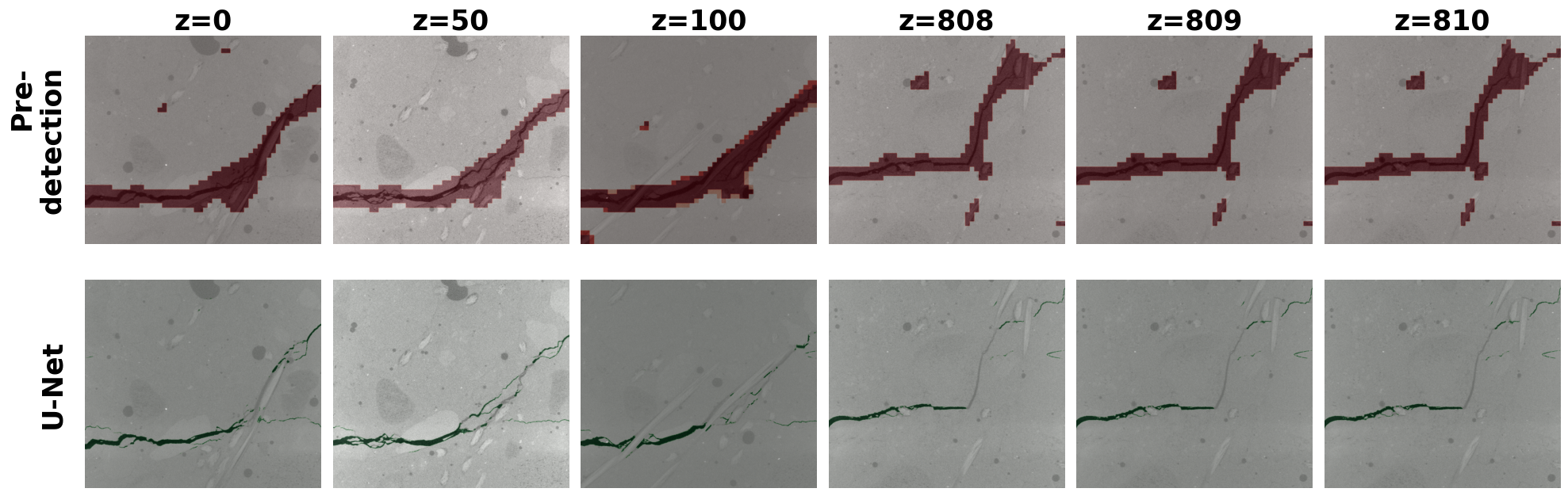}
 	\caption{
 		Comparison between the proposed statistical localization method (top row)
 		and the U-Net segmentation (bottom row) on selected slices of the 
 		real CT volume NC-1. 
 		Each column corresponds to a different depth ($z = 0$, $50$, $100$, $808$, 
 		$809$, $810$). 
 	}
 	\label{fig:methodcompare1}
 \end{figure}

 Figure \ref{fig:methodcompare1} presents a qualitative comparison for the
 NC-1 volume. In this section, In the first three slices of this figure, although the U-Net produces a voxel-wise segmentation mask and the statistical
 method returns the union of detected anomalous cubes, both indicate the
 presence of a crack in these sections. In this section, green denotes the voxel-wise U-Net segmentation. 
In the last three slices of Figure \ref{fig:methodcompare1}, some crack sections are not recognized by the U-Net, whereas the multiple statistical test continues to classify the corresponding neighbourhood as anomalous. Although a few regions are incorrectly detected as anomalous by the statistical procedure, the total misclassified volume remains small compared with the size of the correctly localized cracks.

\begin{figure}[H]
	\centering
	\includegraphics[width=0.65\textwidth]{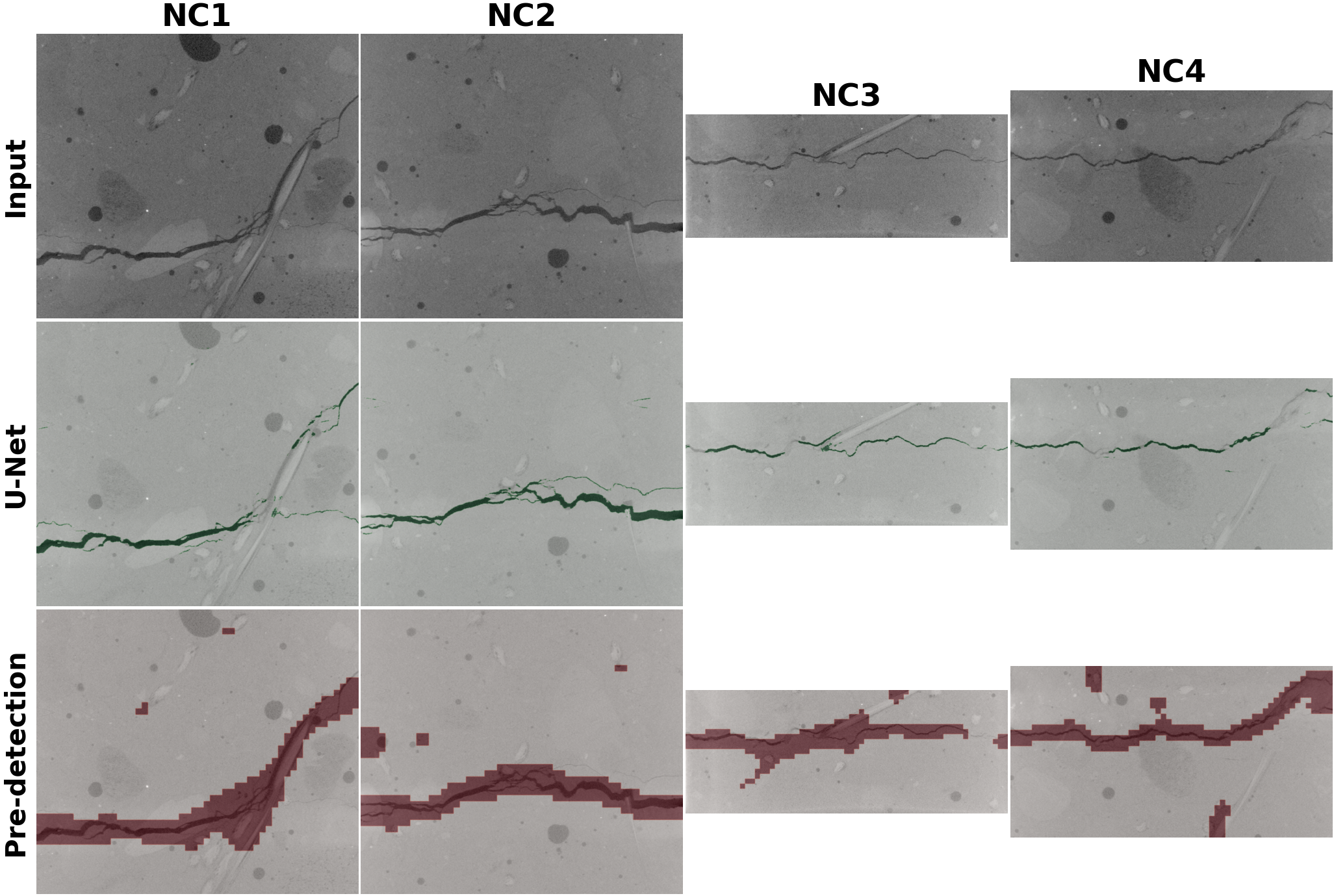}
	\caption{
		Representative crack localization results for all four NC volumes 
		(NC-1 to NC-4).  
		For each scan, a representative slice is shown alongside the 
		corresponding U-Net segmentation and the anomaly map produced by 
		the proposed statistical test.  
		The method successfully identifies the main crack regions with 
		only minor additional detections false caused by texture variations.
	}
	\label{fig:realNC}
\end{figure}

\begin{figure}[ht]
	\centering
	\includegraphics[width=0.65\textwidth]{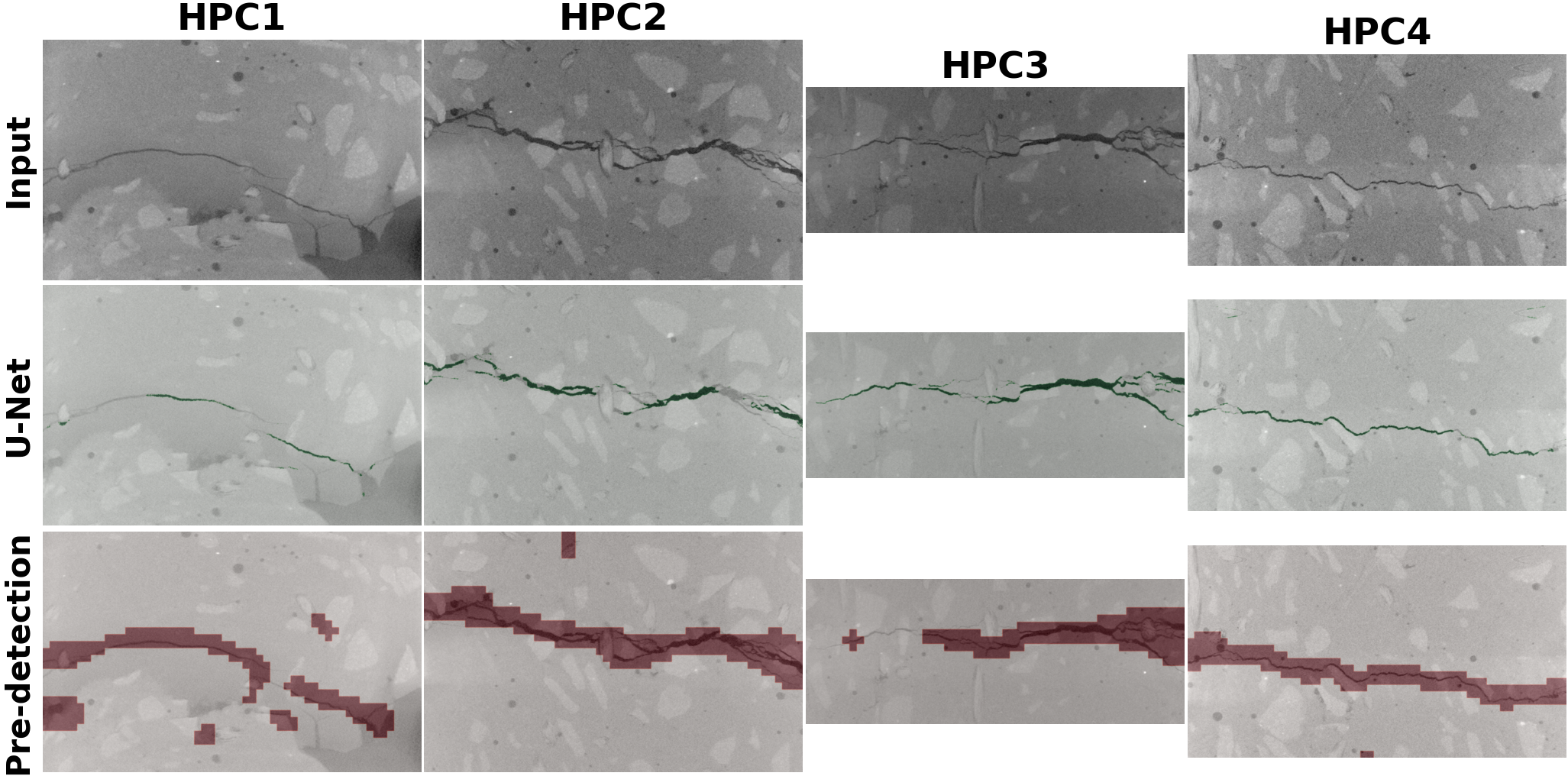}
	\caption{
		Representative crack localization results for the four HPC volumes 
		(HPC-1 to HPC-4).  
		Despite higher material heterogeneity, the localization procedure reliably identifies the 
		regions containing cracks and aligns well with the segmentation 
		produced by the U-Net.  
		Occasional false positives appear along bright elongated grains, but 
		the overall misclassified volume remains small compared with the 
		true crack extent.
	}
	\label{fig:realHPC}
\end{figure}

 To summarise the results of the proposed procedure across the entire set of
 real CT datasets, Figures \ref{fig:realNC} and \ref{fig:realHPC} show representative slices
 for all eight images (NC-1 to NC-4 and HPC-1 to HPC-4). Across both NC and HPC samples, our procedure detects crack regions reliably and covers most of the structures that the deep-learning segmentation highlights.
 

Across both NC and HPC samples, the procedure detects crack regions reliably 
and covers most of the structures highlighted by the deep-learning 
segmentation.  
The results are consistent with the results obtained for the semi-synthetic dataset in Section \ref{subsec:nummericsyn}, where the method performed well across a wide range of concrete types and crack geometries.
On real CT data, the localization generally aligns with the U-Net segmentation, indicating that the statistical test captures the dominant crack geometry even under natural material variability.  
False positives arise occasionally along elongated grains or strong texture transitions because the maximal Hessian entry filter is highly edge-sensitive. However, the overall falsely detected volume remains small relative to the correctly localized crack regions.
 Overall, the localization remains robust across the dataset and reproduces the crack regions with good geometric fidelity.
 
A limitation of the procedure appears in the cases NC-3 and HPC-3 (see Figures \ref{fig:realNC} and \ref{fig:realHPC}), where the cracks are extremely thin and exhibit grey values very close to the homogeneous material. 
By contrast, the U-Net segmentation model, which was trained on the VoroCrack3D dataset containing numerous examples of thin crack (width $w = 1$), preserves these subtle structures and therefore recovers the thin fracture more reliably. This discrepancy highlights an inherent drawback of the current approach: relying solely on the Maximal Hessian entry, which causes the loss of curvature information carried by its other eigenvalues. In these situations, the Maximal Hessian entry filter, is unable to extract stable directional characteristics. Consequently, it remains dominated by noise, making the CUSUM values fail to distinguishable from foreground and background.

\subsection{Run time} \label{subsec:runtime}

All experiments were carried out on an NVIDIA RTX 3090 GPU with 24 GB memory. Since the proposed method performs statistical localization rather than voxel-wise segmentation, its runtime is not intended as a competitive benchmark against the 3D U-Net used in Section \ref{subsec:nummericreal}.

For the semi-synthetic $400^3$ images, the total runtime of our procedure is on the same order of magnitude as the forward pass of the U-Net, typically about 1.3-1.5 times longer. A similar behaviour appears for the real CT scans. For the HPC volumes, whose dimensions range from approximately $640 \times 400 \times 1040$ to $1000 \times 920 \times 1300$ voxels, 
the U-Net processes a scan in roughly $37$-$91$ seconds, 
while our localization procedure requires about $53$-$119$ seconds. 
For the NC scans of similar size, the U\textsc{-}Net runs in approximately 
$52$-$117$ seconds, and our method completes in about $71$-$167$ seconds.
 These ranges show that the statistical localization remains within the same overall time scale as the deep-learning model, even though the two approaches address different tasks. Furthermore, the computational cost of our statistical pipeline increases linearly with the number of voxels, $O(|W|)$, since each Hessian filtering, geometric statistic extraction on fixed-size cubes, and scanning window tests contributes only constant work per voxel or per cube. The GPU implementation ensures that these operations remain efficient on large CT data.

Because the runtimes of both approaches are comparable, it is feasible in practice to apply the two procedures in combination. The statistical test remains sensitive in regions where the segmentation model may miss subtle crack structures, while the U-Net may recover fine crack structures that the statistical procedure is not able to identify. Using both together provides an opportunity to reduce the chance of missing crack regions.

\section{Discussion} \label{sec:discussion}

The experimental results show that the proposed statistical framework provides a reliable and computationally efficient approach for crack pre-localization in large three-dimensional concrete CT images. Across both semi-synthetic and real datasets, the method consistently identifies the main crack regions using only a simple Hessian-based preprocessing step, geometric descriptors on a regular cube partition, and a spatial multiple testing procedure that accounts for dependence among overlapping windows.

On the semi-synthetic volumes, the Maximal Hessian Entry filter proves sufficiently informative to preserve the local geometry of cracks, even though it is noisier than eigenvalue-based alternatives such as Frangi or Sheetness filters. Because the subsequent analysis relies on geometric statistics rather than voxel-level detail, the method does not require a precise segmentation. This explains the stability of the geometric descriptors  across a wide range of crack widths, orientations, and noise levels. The parameter study further indicates that the procedure is robust to the choice of $\alpha$ once the sparsity threshold is set near $\tau \approx 0.10$, confirming that decisions are driven primarily by local geometric structure.

The evaluation on real CT scans supports these findings. The localization results generally align with the U-Net segmentations, despite the differences in methodology. The statistical procedure reacts to abrupt changes in local shape, while the U-Net relies on pre-trained patterns. As a consequence, there are regions where the U-Net detects fine crack structures that the statistical method cannot indicate, and regions where the statistical test highlights anomalies that the U-Net classifies only weakly or misses entirely crack voxels. This complementarity is consistent with the different objectives of each method: the U-Net aims for exact segmentation, whereas the statistical procedure is designed to detect regions with anomalous geometric characteristics. The qualitative comparisons show that the statistical method provides a robust pre-localization signal, with false positives rate controlled.

In real CT data, intensity variations often include speckle-like fluctuations that do not correspond to actual material features. Speckle interference produces small bright or dark spots that behave like isolated pores or noise artifacts rather than crack structures. Such isolated spots can be seen, for example, in the slices NC-1, NC-3, and HPC-2 of Figure \ref{fig:nchpc}. However, as the maximal Hessian entry filter responds only weakly to isolated spots, and the cube-level geometric statistics smooth out their effect, and the subsequent geometric statistics average these effects over fixed-size cubes, preventing individual noisy voxels from influencing the descriptors. When the CUSUM statistic and the $p$-value calculation are applied across overlapping windows, any remaining speckle effects fail to produce a consistent signal. As a result, isolated speckle patterns do not persist in the final localization map.

The main limitations arise when cracks are extremely thin or have grayvalues nearly indistinguishable from the background. In such cases, the maximal Hessian entry becomes dominated by noise, leading to weak geometric contrasts and CUSUM values that do not separate anomalous and homogeneous regions. Similar difficulties arise when cracks are wider than the cube size or cover a large portion of the volume, in which case the geometric features may no longer characterise the region reliably. These issues suggest natural extensions, including multiscale cube constructions, curvature-sensitive filter design, or different combination of geometric statistics.

Overall, the framework provides a predictable and interpretable tool for pre-localizing cracks in both semi-synthetic and real CT data. Its linear computational complexity, robustness to material noise, and compatibility with deep learning models make it well suited for integration into large-scale inspection pipelines.

\section{Conclusions} \label{sec:conl}

This paper presents a statistical framework designed for effective crack pre-localization in large-scale image processing, tackling the computational challenges faced by traditional methods. The proposed method strikes a balance between efficiency and accuracy by employing Hessian-based filters and examining geometric structures within image subregions, as mentioned in Section \ref{subsec:segment}.  

The key feature of this approach is the combination of a simple edge-detection technique with geometric analysis, where the false positive rate is controlled during the final testing. This integration not only enhances the performance of the Maximal Hessian Entry filter but also maintains computational feasibility for detecting cracks in high-resolution images. Results from tests on both semi-synthetic and real 3D CT images demonstrate the method’s effectiveness in accurately localizing cracks within a reasonable timeframe. Once our crack detection algorithm localizes potential image regions containing cracks, the subsequent precise crack segmentation methods (including Deep Learning techniques) will be performed on these subregions, resulting in drastic runtime reduction and enhancing the effectiveness of crack segmentation. Therefore, this is advantageous in that the method can be easily integrated with other workflows and be compatible with deep learning techniques.

A further practical advantage of the framework is its minimal dependence on training data. In the experiments of this paper, all calibration steps rely on a single homogeneous $400^3$ CT volume, which already suffices to estimate the empirical null distribution for the testing procedure. Unlike learning-based segmentation models, the method does not require large annotated datasets or extensive training, making it particularly suitable in settings where labelled crack masks are costly to obtain.

The extended assessment in this study shows that the statistical procedure remains robust across highly heterogeneous real concrete scans and aligns well with the structures detected by a 3D U-Net trained on the VoroCrack3D dataset. While the two methods serve different purposes, their agreement across most regions confirms the stability of the geometric descriptors and the spatial testing framework. In several cases, the U-Net captures fine crack structures that the statistical method cannot resolve, whereas the statistical test highlights additional anomalous areas that the U-Net does not clearly indicate. Combined with the observation that both procedures have comparable runtimes, this suggests that using both methods jointly reduces the likelihood of overlooking relevant crack regions in complex CT images.

Scaling the framework to volumes of size $(10000 \times 10000 \times 2000)$ introduces additional challenges, since such data cannot be processed at once and must be handled in smaller batches. This naturally leads to two possible strategies. A first option is to link the statistical localization with a lightweight segmentation network: the statistical method would identify candidate crack regions, the lightweight model would provide a secondary check, and the full-resolution, computationally expensive model, which is currently under development, would then be applied only to the regions suggested by both methods. This is considerably more feasible than running a heavy network over the entire volume, which would require large GPU resources, substantial training time, extensive hyperparameter tuning, and repeated inspection of false detections.

A second option is to use the localization procedure directly during training of segmentation models. Running our method on a set of training volumes would provide approximate crack regions without the need for manual labeling, and thereby focuses the training loss on structurally informative regions. This would reduce the amount of irrelevant background processed during training and may accelerate convergence, especially for very large datasets. Since non-crack regions dominate large scans, excluding them early is computationally advantageous and remains compatible with a batched training pipeline. Although this direction requires systematic validation, it appears feasible and aligns with the interpretability properties of the statistical features.

Future developments should also aim at improving the preprocessing stage. At present, the method depends entirely on the maximal Hessian entry filter, whose loss of curvature information limits sensitivity to very thin or low-contrast cracks. Designing enhanced second-order filters that balance efficiency with curvature-awareness would therefore increase robustness. Additionally, understanding the spatial correlation structure induced by cracks within the geometric statistics field may allow the construction of sharper test statistics and more accurate local decision rules. These refinements would strengthen the classification step and improve localization quality, particularly in materials with complex or highly heterogeneous texture.

\section*{Funding}
This research was funded by the German Federal Ministry of Research, Technology and Space (BMFTR) [grant number 05M20VUA (DAnoBi)].

\section*{Declaration of Competing Interest}
The authors declare that they have no known competing financial interests or personal relationships that could have appeared to influence the work reported in this paper.

\bibliographystyle{elsarticle-num}
\bibliography{refs.bib}

@article{guo2023,
	author  = {Guo, Y. and Chen, X. and Wang, Z. and Ning, Y. and Bai, L.},
	title   = {Identification of mixed mode damage types on rock-concrete interface under cyclic loading},
	journal = {International Journal of Fatigue},
	volume  = {166},
	pages   = {107273},
	year    = {2023},
	doi     = {10.1016/j.ijfatigue.2022.107273}
}

@inproceedings{ogawa2019,
	author    = {Ogawa, S. and Matsushima, K. and Takahashi, O.},
	title     = {Crack Detection Based on Gaussian Mixture Model using Image Filtering},
	booktitle = {International Symposium on Electrical and Electronics Engineering},
	pages     = {79--84},
	year      = {2019},
	doi       = {10.1109/ISEE2.2019.8921060}
}

@incollection{frangi98,
	author    = {Frangi, A. F. and Niessen, W. J. and Vincken, K. L. and Viergever, M. A.},
	title     = {Multiscale vessel enhancement filtering},
	booktitle = {Lecture Notes in Computer Science},
	volume    = {1496},
	pages     = {130--137},
	year      = {2000},
	doi       = {10.1007/BFb0056195}
}

@article{sato2000tissue,
	author       = {Y. Sato and S. Nakajima and N. Shiraga and H. Atsumi and S. Yoshida and T. Koller and G. Gerig and R. Kikinis},
	title        = {Tissue classification based on {3D} local intensity structures for volume rendering},
	journal      = {IEEE Transactions on Visualization and Computer Graphics},
	volume       = {6},
	number       = {2},
	pages        = {160--180},
	year         = {2000},
	month        = apr,
	doi          = {10.1109/2945.856997},

}

@article{Yamaguchi2010,
	author  = {Yamaguchi, T. and Hashimoto, S.},
	title   = {Fast crack detection method for large-size concrete surface images using percolation-based image processing},
	journal = {Machine Vision and Applications},
	volume  = {21},
	number  = {5},
	pages   = {797--809},
	year    = {2010},
	doi     = {10.1007/s00138-009-0189-8}
}

@inproceedings{ehrig2011comparison,
	author    = {Ehrig, K. and Goebbels, J. and Meinel, D. and Paetsch, O. and Prohaska, S. and Zobel, V.},
	title     = {Comparison of Crack Detection Methods for Analyzing Damage Processes in Concrete with Computed Tomography},
	booktitle = {International Symposium on Digital Industrial Radiology and Computed Tomography},
	year      = {2011},
	url       = {https://www.ndt.net/?id=11150}
}

@article{barisin2022,
	author  = {Barisin, T. and Jung, C. and Müsebeck, F. and Redenbach, C. and Schladitz, K.},
	title   = {Methods for segmenting cracks in 3D images of concrete: A comparison based on semi-synthetic images},
	journal = {Pattern Recognition},
	volume  = {129},
	year    = {2022},
	doi     = {10.1016/j.patcog.2022.108747}
}

@article{JUNG2024110474,
	author  = {Jung, C. and Redenbach, C. and Schladitz, K.},
	title   = {VoroCrack3d: An annotated semi-synthetic 3d image data set of cracked concrete},
	journal = {Data in Brief},
	volume  = {54},
	pages   = {110474},
	year    = {2024},
	doi     = {10.1016/j.dib.2024.110474}
}

@book{brodsky1993nonparametric,
	author    = {Brodsky, B. E. and Darkhovsky, B. S.},
	title     = {Nonparametric Methods in Change Point Problems},
	publisher = {Springer Dordrecht},
	year      = {1993}
}

@article{Alonso-Ruiz_Spodarev_2017,
	author  = {Alonso-Ruiz, P. and Spodarev, E.},
	title   = {Estimation of entropy for Poisson marked point processes},
	journal = {Advances in Applied Probability},
	volume  = {49},
	pages   = {258--278},
	year    = {2017},
	url     = {http://www.jstor.org/stable/44985422}
}

@article{AlonsoRuiz_Spodarev_2018,
	author  = {Alonso Ruiz, P. and Spodarev, E.},
	title   = {Entropy-based Inhomogeneity Detection in Fiber Materials},
	journal = {Methodology and Computing in Applied Probability},
	volume  = {20},
	pages   = {1223--1239},
	year    = {2018},
	doi     = {10.1007/s11009-017-9603-2}
}

@article{Dresvyanskiy2020,
	author  = {Dresvyanskiy, D. and Karaseva, T. and Makogin, V. and Mitrofanov, S. and Redenbach, C. and Spodarev, E.},
	title   = {Detecting anomalies in fibre systems using 3-dimensional image data},
	journal = {Statistical Computing},
	volume  = {30},
	pages   = {817--837},
	year    = {2020},
	doi     = {10.1007/s11222-020-09921-1}
}

@article{sidak67,
	author  = {Šidák, Z.},
	title   = {Rectangular Confidence Regions for the Means of Multivariate Normal Distributions},
	journal = {Journal of the American Statistical Association},
	volume  = {62},
	number  = {318},
	pages   = {626--633},
	year    = {1967},
	doi     = {10.1080/01621459.1967.10482935}
}

@article{hochberg1988,
	author  = {Hochberg, Y.},
	title   = {A sharper {B}onferroni procedure for multiple tests of significance},
	journal = {Biometrika},
	volume  = {75},
	number  = {4},
	pages   = {800--802},
	year    = {1988},
	doi     = {10.1093/biomet/75.4.800}
}

@article{holm79,
	author  = {Holm, S.},
	title   = {A Simple Sequentially Rejective Multiple Test Procedure},
	journal = {Scandinavian Journal of Statistics},
	volume  = {6},
	number  = {2},
	pages   = {65--70},
	year    = {1979},
	url     = {http://www.jstor.org/stable/4615733}
}

@article{Benjamini1995,
	author  = {Benjamini, Y. and Hochberg, Y.},
	title   = {Controlling the False Discovery Rate: A Practical and Powerful Approach to Multiple Testing},
	journal = {Journal of the Royal Statistical Society: Series B},
	volume  = {57},
	number  = {1},
	pages   = {289--300},
	year    = {1995},
	doi     = {10.1111/j.2517-6161.1995.tb02031.x}
}

@article{cai2021,
	author  = {Cai, T. T. and Sun, W. and Xia, Y.},
	title   = {{LAWS}: A locally adaptive weighting and screening approach to spatial multiple testing},
	journal = {Journal of the American Statistical Association},
	volume  = {117},
	number  = {539},
	pages   = {1370--1383},
	year    = {2021},
	doi     = {10.1080/01621459.2020.1859379}
}

@article{lei2018,
	author  = {Lei, L. and Fithian, W.},
	title   = {AdaPT: An Interactive Procedure for Multiple Testing with Side Information},
	journal = {Journal of the Royal Statistical Society: Series B},
	volume  = {80},
	pages   = {649--679},
	year    = {2018},
	doi     = {10.1111/rssb.12274}
}

@article{lit2019,
	author  = {Li, A. and Barber, R. F.},
	title   = {Multiple Testing with the Structure-Adaptive {B}enjamini-{H}ochberg Algorithm },
	journal = {Journal of the Royal Statistical Society: Series B},
	volume  = {81},
	pages   = {45--74},
	year    = {2019},
	doi     = {10.1111/rssb.12298}
}

@book{ohser20093d,
	author    = {Ohser, J. and Schladitz, K.},
	title     = {3D images of materials structures: processing and analysis},
	publisher = {John Wiley \& Sons},
	year      = {2009}
}

@book{tartakovsky2014sequential,
	author    = {Tartakovsky, A. and Nikiforov, I. and Basseville, M.},
	title     = {Sequential Analysis: Hypothesis Testing and Changepoint Detection},
	publisher = {CRC Press},
	year      = {2014}
}

@article{annika22,
	author  = {Betken, A. and Wendler, M.},
	title   = {Rank-based change-point analysis for long-range dependent time series},
	journal = {Bernoulli},
	volume  = {28},
	number  = {4},
	pages   = {2209--2233},
	year    = {2022},
	doi     = {10.3150/21-BEJ1416}
}

@InProceedings{pmlr-v97-ghorbani19b,
	title = 	 {An Investigation into Neural Net Optimization via Hessian Eigenvalue Density},
	author =       {Ghorbani, Behrooz and Krishnan, Shankar and Xiao, Ying},
	booktitle = 	 {Proceedings of the 36th International Conference on Machine Learning},
	pages = 	 {2232--2241},
	year = 	 {2019},
	editor = 	 {Chaudhuri, Kamalika and Salakhutdinov, Ruslan},
	volume = 	 {97},
	series = 	 {Proceedings of Machine Learning Research},
	month = 	 {09--15 Jun},
	publisher =    {PMLR},
	url = 	 {https://proceedings.mlr.press/v97/ghorbani19b.html},

}

@article{orlowski2009efficient,
	author       = {P. Or{\l}owski and M. Orkisz},
	title        = {Efficient computation of Hessian-based enhancement filters for tubular structures in {3D} images},
	journal      = {IRBM},
	volume       = {30},
	number       = {3},
	pages        = {128--132},
	year         = {2009},
	issn         = {1959-0318},
	doi          = {10.1016/j.irbm.2009.04.003},

}

@article{hare2023detecting,
	author       = {W. Hare and C. W. Royer},
	title        = {Detecting negative eigenvalues of exact and approximate Hessian matrices in optimization},
	journal      = {Optimization Letters},
	volume       = {17},
	number       = {},
	pages        = {1739--1756},
	year         = {2023},
	doi          = {10.1007/s11590-023-02033-5},

}

@article{yang2014fast,
	author       = {Shih{-}Feng Yang and Ching{-}Hsue Cheng},
	title        = {Fast computation of Hessian-based enhancement filters for medical images},
	journal      = {Computer Methods and Programs in Biomedicine},
	volume       = {116},
	number       = {3},
	pages        = {215--225},
	year         = {2014},
	issn         = {0169-2607},
	doi          = {10.1016/j.cmpb.2014.05.002},

}

@article{jerman2016enhancement,
	author       = {T. Jerman and F. Pernu{\v{s}} and B. Likar and {\v{Z}}. {\v{S}}piclin},
	title        = {Enhancement of Vascular Structures in {3D} and {2D} Angiographic Images},
	journal      = {IEEE Transactions on Medical Imaging},
	volume       = {35},
	number       = {9},
	pages        = {2107--2118},
	year         = {2016},
	month        = sep,
	doi          = {10.1109/TMI.2016.2550102},

}

@inproceedings{salamon2025gulliver,
	author       = {M. Salamon and N. Reims and D. Prjamkov and M. Schmitt and J. Makarov and D. Ak and M. Kronenberger and K. Schladitz and C. Redenbach and S. Grzesiak and C. de Sousa and C. Thiele and M. Pahn},
	title        = {Gulliver -- A new kind of industrial {CT}},
	booktitle    = {14th Conference on Industrial Computed Tomography (iCT)},
	address      = {Antwerp, Belgium},
	month        = feb,
	year         = {2025},
	pages        = {},
	publisher    = {e-Journal of Nondestructive Testing, Vol. 30(2)},
	doi          = {10.58286/30741},

}

@inproceedings{nowacka2024deep,
	author       = {A. Nowacka and C. Jung and K. Schladitz and C. Redenbach and M. Pahn},
	title        = {Deep learning models for crack segmentation in {3D} images of concrete trained on semi-synthetic data},
	booktitle    = {13th Conference on Industrial Computed Tomography (iCT) 2023},
	address      = {Wels, Austria},
	month        = feb,
	year         = {2024},
	publisher    = {e-Journal of Nondestructive Testing, Vol. 29(3)},
	doi          = {10.58286/29241},

}

@article{kaveh2024recent,
	author       = {H. Kaveh and R. Alhajj},
	title        = {Recent advances in crack detection technologies for structures: a survey of 2022--2023 literature},
	journal      = {Frontiers in Built Environment},
	volume       = {10},
	pages        = {1321634},
	year         = {2024},
	doi          = {10.3389/fbuil.2024.1321634},

}

@article{he2025quantitative,
	author       = {Jialuo He and Yong Deng and Xianming Shi},
	title        = {Quantitative analysis of pore structures and microcracks in self-healing concrete after freeze-thaw exposure: An X-ray computed tomography-based approach},
	journal      = {Cement and Concrete Composites},
	volume       = {162},
	pages        = {106105},
	year         = {2025},
	issn         = {0958-9465},
	doi          = {10.1016/j.cemconcomp.2025.106105},

}

@article{qi2025crack,
	author       = {Yahui Qi and Pengzhen Lin and Guojun Yang and Tao Liang},
	title        = {Crack detection and {3D} visualization of crack distribution for {UAV}-based bridge inspection using efficient approaches},
	journal      = {Structures},
	volume       = {78},
	pages        = {109075},
	year         = {2025},
	issn         = {2352-0124},
	doi          = {10.1016/j.istruc.2025.109075},

}

@article{jung2023crack,
	author       = {C. Jung and C. Redenbach},
	title        = {Crack modeling via minimum-weight surfaces in {3D} Voronoi diagrams},
	journal      = {Journal of Mathematics in Industry},
	volume       = {13},
	pages        = {10},
	year         = {2023},
	doi          = {10.1186/s13362-023-00138-1},

}

@article{xu2025crack,
	author       = {Jiangbo Xu and Shaowei Wang and Ruida Han and Xiong Wu and Danni Zhao and Xianglong Zeng and Ruibo Yin and Zemin Han and Yifan Liu and Sheng Shu},
	title        = {Crack segmentation and quantification in concrete structures using a lightweight {YOLO} model based on pruning and knowledge distillation},
	journal      = {Expert Systems with Applications},
	volume       = {283},
	pages        = {127834},
	year         = {2025},
	issn         = {0957-4174},
	doi          = {10.1016/j.eswa.2025.127834},

}

@article{yu2022cracklab,
	author       = {Zhenwei Yu and Yonggang Shen and Zhilin Sun and Jiang Chen and Wu Gang},
	title        = {Cracklab: A high-precision and efficient concrete crack segmentation and quantification network},
	journal      = {Developments in the Built Environment},
	volume       = {12},
	pages        = {100088},
	year         = {2022},
	issn         = {2666-1659},
	doi          = {10.1016/j.dibe.2022.100088},

}

@article{zhou2025mpa,
	author       = {Yanli Zhou and Zhanfang Zhao},
	title        = {{MPA-YOLO}: Steel surface defect detection based on improved {YOLO}v8 framework},
	journal      = {Pattern Recognition},
	volume       = {168},
	pages        = {111897},
	year         = {2025},
	issn         = {0031-3203},
	doi          = {10.1016/j.patcog.2025.111897},

}

@article{li2025high,
	author       = {Shufang Li and Xiaolin Yao and Najah M. L. Al Maimuri and Elimam Ali and H. Elhosiny Ali and H. Aslza and Jos{\'e} Escorcia{-}Gutierrez},
	title        = {High quality training set development for road crack detection using progressive data-driven models with integrated identification},
	journal      = {Case Studies in Construction Materials},
	volume       = {23},
	pages        = {e05416},
	year         = {2025},
	issn         = {2214-5095},
	doi          = {10.1016/j.cscm.2025.e05416},
	url          = {https://doi.org/10.1016/j.cscm.2025.e05416}
}

@article{xie2025versatile,
	author       = {Jiawei Xie and Baolin Chen and Anna Giacomini and Hongyu Guo and Umair Iqbal and Jinsong Huang},
	title        = {A versatile synthetic data generation framework for crack detection},
	journal      = {Engineering Structures},
	volume       = {344},
	pages        = {121428},
	year         = {2025},
	issn         = {0141-0296},
	doi          = {10.1016/j.engstruct.2025.121428},

}

@article{xia2026image,
	author       = {Changqing Xia and Hao Chen and Min Zhu and Bowen Lin and Xiangsheng Chen},
	title        = {Image super-resolution reconstruction based on conditional diffusion model in crack detection and segmentation of shield tunnels},
	journal      = {Automation in Construction},
	volume       = {181},
	pages        = {106570},
	year         = {2026},
	issn         = {0926-5805},
	doi          = {10.1016/j.autcon.2025.106570},

}

@article{morozova2022visualization,
	author       = {N. Morozova and K. Shibano and Y. Shimamoto and S. Tayfur and N. Alver and T. Suzuki},
	title        = {Visualization and evaluation of concrete damage in-service headworks by X-ray {CT} and non-destructive inspection methods},
	journal      = {Frontiers in Built Environment},
	volume       = {8},
	pages        = {947759},
	year         = {2022},
	doi          = {10.3389/fbuil.2022.947759},

}

@article{avendano2024image,
	author       = {J. C. Avenda{\~n}o and J. Leander and R. Karoumi},
	title        = {Image-Based Concrete Crack Detection Method Using the Median Absolute Deviation},
	journal      = {Sensors},
	volume       = {24},
	number       = {9},
	pages        = {2736},
	year         = {2024},
	doi          = {10.3390/s24092736},

}

@article{zawad2021comparative,
	author       = {Md. Rahat Zawad and Md. Fahad Zawad and Md. Asifur Rahman and Sudipto Priyom},
	title        = {A Comparative Review of Image Processing Based Crack Detection Techniques on Civil Engineering Structures},
	journal      = {Journal of Soft Computing in Civil Engineering},
	volume       = {5},
	number       = {3},
	pages        = {58--74},
	year         = {2021},
	publisher    = {Pouyan Press},
	issn         = {2588-2872},
	doi          = {10.22115/scce.2021.287729.1325},

}

@article{lee2024parametric,
	author       = {Dong{-}Eun Lee and Young Choi and Geuntae Hong and M. Maruthi and Chang{-}Yong Yi and Young{-}Jun Park},
	title        = {Parametric image-based concrete defect assessment method},
	journal      = {Case Studies in Construction Materials},
	volume       = {20},
	pages        = {e02962},
	year         = {2024},
	issn         = {2214-5095},
	doi          = {10.1016/j.cscm.2024.e02962},

}

@article{jia2022experimental,
	author       = {Mengdi Jia and Zhimin Wu and Rena C. Yu and Xiaoxin Zhang},
	title        = {Experimental investigation of mixed mode {I--II} fatigue crack propagation in concrete using a digital image correlation method},
	journal      = {Engineering Fracture Mechanics},
	volume       = {272},
	pages        = {108712},
	year         = {2022},
	issn         = {0013-7944},
	doi          = {10.1016/j.engfracmech.2022.108712},

}

@article{flah2020classification,
	author       = {Majdi Flah and Ahmed R. Suleiman and Moncef L. Nehdi},
	title        = {Classification and quantification of cracks in concrete structures using deep learning image-based techniques},
	journal      = {Cement and Concrete Composites},
	volume       = {114},
	pages        = {103781},
	year         = {2020},
	issn         = {0958-9465},
	doi          = {10.1016/j.cemconcomp.2020.103781},
	url          = {https://doi.org/10.1016/j.cemconcomp.2020.103781}
}

@article{kang2022efficient,
	author       = {D. H. Kang and Y. J. Cha},
	title        = {Efficient attention-based deep encoder and decoder for automatic crack segmentation},
	journal      = {Structural Health Monitoring},
	volume       = {21},
	number       = {5},
	pages        = {2190--2205},
	year         = {2022},
	doi          = {10.1177/14759217211053776},

}

@article{zhu2024lightweight,
	author       = {A. Zhu and J. Xie and B. Wang and others},
	title        = {Lightweight defect detection algorithm of tunnel lining based on knowledge distillation},
	journal      = {Scientific Reports},
	volume       = {14},
	pages        = {27178},
	year         = {2024},
	doi          = {10.1038/s41598-024-77404-8},

}

@article{morozova2023frost,
	author       = {Nadezhda Morozova and Kazuma Shibano and Yuma Shimamoto and Sena Tayfur and Ninel Alver and Tetsuya Suzuki},
	title        = {Frost damage evaluation of concrete irrigation structure by X-ray {CT} and {AE} energy release trend at the initial loading stage},
	journal      = {Case Studies in Construction Materials},
	volume       = {18},
	pages        = {e02088},
	year         = {2023},
	issn         = {2214-5095},
	doi          = {10.1016/j.cscm.2023.e02088},

}

@book{ramsay2005fda,
	author       = {J. O. Ramsay and B. W. Silverman},
	title        = {Functional Data Analysis},
	edition      = {2},
	series       = {Springer Series in Statistics},
	publisher    = {Springer},
	address      = {New York, NY},
	year         = {2005},
	isbn         = {978-0-387-40080-8},
	doi          = {10.1007/b98888},

}

@article{liebl2023fast,
	author       = {Dominik Liebl and Matthew Reimherr},
	title        = {Fast and fair simultaneous confidence bands for functional parameters},
	journal      = {Journal of the Royal Statistical Society: Series B (Statistical Methodology)},
	volume       = {85},
	number       = {3},
	pages        = {842--868},
	year         = {2023},
	doi          = {10.1093/jrsssb/qkad026},
	issn         = {1369-7412},

}

@article{sarkar2022local,
	author       = {Sanat K. Sarkar and Zhigen Zhao},
	title        = {Local false discovery rate based methods for multiple testing of one-way classified hypotheses},
	journal      = {Electronic Journal of Statistics},
	volume       = {16},
	number       = {2},
	pages        = {6043--6085},
	year         = {2022},
	doi          = {10.1214/22-EJS2080},

}

@article{cao2022optimal,
	author       = {Hongyuan Cao and Jun Chen and Xianyang Zhang},
	title        = {Optimal false discovery rate control for large scale multiple testing with auxiliary information},
	journal      = {Annals of Statistics},
	volume       = {50},
	number       = {2},
	pages        = {807--857},
	year         = {2022},
	doi          = {10.1214/21-AOS2128},

}

\end{document}